\documentclass[twoside,11pt]{article}

\usepackage{jmlr2e}

\usepackage{amsmath}
\usepackage{amsfonts}
\usepackage{mathrsfs}
\usepackage{color}
\usepackage[titletoc]{appendix}
\usepackage{bm}
\usepackage{algorithm}
\usepackage{algpseudocode}
\usepackage{algpascal}
\usepackage{url}
\usepackage{afterpage}
\usepackage{enumitem} 
\usepackage{multirow}
\usepackage{booktabs}
\usepackage{subfigure}

\numberwithin{equation}{section}

\def\cB{\mathcal B}
\def\cC{\mathcal C}

\def\cF{\mathcal F}
\def\cG{\mathcal G}
\def\cH{\mathcal H}

\def\cJ{\mathcal J}

\def\cM{\mathcal M}
\def\cN{\mathcal N}
\def\cP{\mathcal P}

\def\cY{\mathcal Y}
\def\cZ{\mathcal Z}

\newcommand{\bA}{{\bf A}}

\newcommand{\bd}{{\bf d}}

\newcommand{\bff}{{\bf f}}
\newcommand{\bg}{{\bf g}}

\newcommand{\bh}{{\bf h}}

\newcommand{\bp}{{\bf p}}

\newcommand{\bt}{{\bf t}}

\newcommand{\bu}{{\bf u}}

\newcommand{\bv}{{\bf v}}

\newcommand{\bx}{{\bf x}}
\newcommand{\bX}{{\bf X}}
\newcommand{\by}{{\bf y}}
\newcommand{\bY}{{\bf Y}}
\newcommand{\bz}{{\bf z}}
\newcommand{\bZ}{{\bf Z}}

\newcommand{\bbI}{{\mathbb I}}
\newcommand{\bbN}{{\mathbb N}}

\newcommand{\bbQ}{{\mathbb Q}}
\newcommand{\bbR}{{\mathbb R}}

\newcommand{\E}{\mathbb{E}}

\newcommand{\bbeta}{\bm{\beta}}

\newcommand{\bepsilon}{\bm{\epsilon}}

\newcommand{\bSigma}{\bm{\Sigma}}

\newcommand{\bmu}{\bm{\mu}}
\newcommand{\bvarphi}{\bm{\varphi}}

\newcommand{\cf}{{\it cf.}}
\newcommand{\eg}{{\it e.g.}}

\newcommand{\iid}{{i.i.d.}}
\newcommand{\Frechet}{Fr\'{e}chet}

\newcommand{\Holder}{H\"{o}lder }

\newcommand{\bc}{\begin{center}}
\newcommand{\ec}{\end{center}}
\newcommand{\be}{\begin{equation}}
\newcommand{\ee}{\end{equation}}
\newcommand{\ba}{\begin{array}}
\newcommand{\ea}{\end{array}}
\newcommand{\bean}{\setlength\arraycolsep{1pt}\begin{eqnarray*}}
\newcommand{\eean}{\end{eqnarray*}}
\newcommand{\bea}{\setlength\arraycolsep{1pt}\begin{eqnarray}}
\newcommand{\eea}{\end{eqnarray}}
\newcommand{\ben}{\begin{enumerate}}
\newcommand{\een}{\end{enumerate}}
\newcommand{\bed}{\begin{itemize}}
\newcommand{\eed}{\end{itemize}}

\DeclareMathOperator*{\argmax}{argmax}

\def\defeq{ \stackrel{\rm def}{=} }

\newcommand{\bzero}{{\bf 0}}

\newcommand{\Id}{\bbI}

\firstpageno{1}

\usepackage{lastpage}
\jmlrheading{24}{2023}{1-\pageref{LastPage}}{9/21; Revised
2/23}{3/23}{21-1099}{Minwoo Chae, Dongha Kim, Yongdai Kim and Lizhen Lin}
\ShortHeadings{A likelihood approach to deep generative models}{Chae, Kim, Kim and Lin}

\begin{document}

\title{A Likelihood Approach to Nonparametric Estimation of a Singular Distribution Using Deep Generative Models}

\author{\name Minwoo Chae \email mchae@postech.ac.kr \\
       \addr Department of Industrial and Management Engineering\\
       Pohang University of Science and Technology\\
       Pohang, Gyeongbuk 37673, South Korea
       \AND
       \name Dongha Kim \email dongha0718@sungshin.ac.kr \\
       \addr School of Mathematics, Statistics and Data Science\\Data Science Center\\
       Sungshin Women's University\\
       Seoul, 02844, South Korea
       \AND
       \name Yongdai Kim \email ydkim0903@gmail.com \\
       \addr Department of Statistics\\
       Seoul National University\\
       Seoul, 08826, South Korea
       \AND
       \name Lizhen Lin \email lizhen.lin@nd.edu \\
       \addr Department of Applied and Computational Mathematics and Statistics\\
       University of Notre Dame\\
       South Bend, IN 46556, USA
       }


\editor{Daniel Roy}

\maketitle

\begin{abstract}
We investigate statistical properties of a likelihood approach to nonparametric estimation of a singular distribution using deep generative models. More specifically, a deep generative model is used to model high-dimensional data that are assumed to concentrate around some low-dimensional structure. Estimating the distribution supported on this low-dimensional structure, such as a low-dimensional manifold, is challenging due to its singularity with respect to the Lebesgue measure in the ambient space. In the considered model, a usual likelihood approach can fail to estimate the target distribution consistently due to the singularity. We prove that a novel and effective solution exists by perturbing the data with an instance noise, which leads to consistent estimation of the underlying distribution with desirable convergence rates. We also characterize the class of distributions that can be efficiently estimated via deep generative models. This class is sufficiently general to contain various structured distributions such as product distributions, classically smooth distributions and distributions supported on a low-dimensional manifold. Our analysis provides some insights on how deep generative models can avoid the curse of dimensionality for nonparametric distribution estimation. We conduct a thorough simulation study and real data analysis to empirically demonstrate that the proposed data perturbation technique improves the estimation performance significantly.
\end{abstract}


\section{Introduction}

Suppose that we have observations $\bX_1, \ldots, \bX_n$ which are \iid\ copies of a $D$-dimensional random vector $\bX$ following the distribution $P_*$.
Without any structural assumption, the problem of estimating $P_*$ or related quantities (\eg\ density, support, etc.) with large dimension $D$ is prohibitively difficult, which is widely known as the curse of dimensionality. To avoid the curse of dimensionality, it is natural to assume that the data locate around some lower-dimensional structure which can be captured by the model $\bX = \bY + \bepsilon$, where $\bY$ is a random vector possessing a specific low-dimensional structure and $\bepsilon$ is a full-dimensional noise vector with small variance.
As an example of low-dimensional structures, one may assume that there exists a low-dimensional manifold on which the probability mass of $\bY$ is concentrated.
For this model, our primary interests are in estimating $Q_*,$ the distribution of $\bY,$ or related quantities.  There is a large literature on estimating the support of $Q_*$, i.e., manifold estimation, see, e.g., \cite{ozakin2009submanifold, puchkin2022structure, genovese2012minimax, genovese2012manifold} and  references therein.  The problem of estimating $Q_*$ on the other hand is much less studied and in general a more challenging problem due to the singularity of $Q_*$ with respect to the Lebesgue measure in the ambient space. \cite{berenfeld2019density} and \cite{ozakin2009submanifold} considered  kernel density estimators for estimating the (Hausdorff) density of $Q_*$ when the data are assumed to be supported on the image of a submanifold embedded in a higher dimensional space, thus no noise is considered.

In this paper, we consider a special form of $\bX = \bY + \bepsilon$,  so-called a probabilistic generative model, which 
models the observation as $\bX = \bff(\bZ) + \bepsilon$, where $\bZ$ and $\bepsilon$ are independent random vectors which are not directly observable. 
The latent variable $\bZ$ is a $d$-dimensional random vector drawn from some known distribution $P_Z$, such as the standard normal or uniform distributions supported on  $\cZ$, an open subset of $\bbR^d$, and $\bff: \cZ \to \bbR^D$ is an unknown function which is often called the \textit{generator} or \textit{generating function}.
The noise vector $\bepsilon$ is assumed to follow the normal distribution $\cN(\bzero_D, \sigma^2 \Id_D)$, where $\bzero_D$ and $\Id_D$ denote the $D$-dimensional zero vector and identity matrix, respectively.
We consider the case of $d < D$, in which the distribution of $\bff(\bZ)$ is singular with respect to the Lebesgue measure on $\bbR^D.$ 

The model $\bX = \bff(\bZ) + \bepsilon$ has been investigated in statistical literature with the name of a nonlinear factor model (\citealp{yalcin2001nonlinear}).
In this paper, we model $\bff$ using deep neural networks (DNNs), which  are known to enjoy universal approximations results (\citealp{cybenko1989approximation, hornik1989multilayer, hornik1990universal}).
Accordingly, we adopt the terminology of a \emph{deep generative model}.
In a deep generative model, instead of directly estimating  $P_*$ or $Q_*$, one may  first construct an estimator $\hat\bff$ and the resulting distribution of $\hat\bff(\bZ)$ will serve as an estimator of $Q_*$. Although this approach does not provide an explicit estimator of $Q_*$, it is easy to draw samples from the estimated distribution.


In recent years, deep generative models have achieved tremendous success for modeling high-dimensional data such as images and videos.
Two  popular approaches are used in practice to construct an estimator $\hat\bff$.
The first one is likelihood-based. Variational approaches (\citealp{kingma2013auto, rezende2014stochastic}) and EM-based algorithms (\citealp{burda2016importance, kim2020casting}) are two most representative learning methods in this class. 
The second approach uses the integral probability metrics (IPM; \citealp{muller1997integral}), often called the adversarial losses in deep learning communities, and constructs an estimator by minimizing these metrics. This approach is widely known as the generative adversarial networks (GAN), originally developed by \cite{goodfellow2014generative} and then generalized in \cite{mroueh2017sobolev, li2017mmd} and \cite{ arjovsky2017wasserstein}, to name a few.

In this work, we focus on the likelihood-based approach and study statistical properties of a sieve maximum likelihood estimator (MLE) of deep generative models under the assumption that $P_*$ is the distribution of $\bX=\bff_*(\bZ)+\bepsilon_*$ for some function $\bff_*:\cZ \to \bbR^D$ and $\bepsilon_*\sim \cN(0, \sigma_*^2 \Id_D)$, where $\sigma_* \geq 0$. The primary goal is to estimate $Q_*$,  the distribution of $\bff_*(\bZ)$ induced from the distribution of $\bZ$ via the true generator $\bff_*$. We obtain several important results for this model.

Firstly, we derive a convergence rate of $\hat Q = Q_{\hat\bff}$ to $Q_*=Q_{\bff_*}$ in terms of the Wasserstein metric (\citealp{villani2003topics}), where $\hat\bff$ is a sieve MLE of $\bff_*$ and $Q_\bff$ denotes the distribution of $\bff(\bZ)$, \cf\ Corollary \ref{cor:rate-smooth} and Theorem \ref{thm:rate-W}.
The convergence rate depends on the noise level $\sigma_*$, intrinsic dimension and smoothness of $\bff_*$; see Section \ref{sec:asymp} for the definition.
More interestingly, Corollary \ref{cor:rate-smooth} and Theorem \ref{thm:rate-W} do not guarantee the consistency of a sieve MLE for very small $\sigma_*$.
To resolve this issue and improve the convergence rate, we propose a novel method to perturb the data.
That is, we obtain a sieve MLE of $\bff_*$ based on the perturbed observation $\widetilde\bX_i = \bX_i + \widetilde\bepsilon_i$, where $\widetilde\bepsilon_i$ is an artificial noise vector following the  distribution $\cN(\bzero_D, \widetilde\sigma^2 \Id_D)$. 
The perturbation level $\widetilde\sigma$ will be chosen carefully to provide a desirable convergence rate.
Note that $\widetilde\bX_i$ always possesses a Lebesgue density $\widetilde p_*$ even when $\sigma_*=0.$
Under general conditions, we derive the convergence rate of a sieve MLE for estimating $\widetilde p_*$ with respect to the Hellinger metric, \cf\ Theorem \ref{thm:rate-general} and Corollary \ref{cor:composition}. 
Then, we derive a Wasserstein convergence rate of a sieve MLE of $Q_*$ based on perturbed observations, \cf\ Theorem \ref{thm:W-rate}.  
Specifically, we attain the convergence rate $\widetilde\epsilon_n + \widetilde\sigma_*$ up to a logarithmic factor, where $\widetilde\epsilon_n$ is the Hellinger convergence rate of the sieve MLE of $\widetilde p_*$, and $\widetilde\sigma_* = \sigma_* + \widetilde \sigma$.
Note that $\widetilde\epsilon_n$ decreases as $\widetilde \sigma$ increases because $\widetilde p_*$ becomes smoother while $\widetilde\sigma_*$ increases.
Hence, the degree of perturbation $\widetilde \sigma$ can be determined by minimizing $\widetilde\epsilon_n + \widetilde\sigma_*$.

Recently, successful cases of data perturbation for learning deep generative models have been reported in \cite{song2019generative, meng2021improved}. 
However, theoretical understanding of the data perturbation is still lacking.
Our results in this paper can provide a theoretical justification for the success of various data perturbation procedures for deep generative models.
Note that most existing theories on deep generative models consider GAN, for which additional noise does not help.

Main results concerning the convergence rates are stated non-asymptotically in the sense that for any fixed $n \geq 1$, we provide sufficient conditions under which certain probabilistic inequalities hold.
Besides the convergence rate of a sieve MLE, we characterize a class of distributions that can be represented by $\bff_*(\bZ)$ for some $\bff_*$.
The class is large enough to include various distributions such as product distributions, classically smooth distributions and distributions supported on a low-dimensional manifold.
As an illustrating example, a class of product distributions has the intrinsic dimension 1, and corresponds to the generalized additive model in the regression setting.
This kind of structure has not been studied in an unsupervised learning framework.
The regularity theory of the optimal transport plays an important role for this characterization.

There are a lot of recent articles studying the statistical properties of the GAN estimator; see Section \ref{ssec:review} for review.
It is a critical limitation of most theoretical studies that they assumed the existence of the smooth Lebesgue density $p_*$ of the underlying distribution $P_*$.
They view the GAN in a nonparametric density estimation framework; the convergence rate directly depends on $D$ and the smoothness level of $p_*$.
Consequently, these results only guarantee that GAN performs as good as classical nonparametric density estimators, and cannot explain why and how it outperforms other methods.
Some recent articles reviewed in Section \ref{ssec:review} go beyond the density estimation framework, but their theories are not exhaustive and possess certain limitations.
In this sense, our results about the convergence rates of a sieve MLE with perturbed data are new and important contributions for deep generative models.
In contrast, the idea of using perturbed data with the GAN estimator has been shown to be ineffective through numerical studies in Section 5, as demonstrated by Figures \ref{fig:gen_samples_baselines_mnist} and \ref{fig:gen_samples_baselines_omniglot}.

Our convergence rate depends on not only the intrinsic dimension of the manifold $\bff_*(\cZ)$, which is much smaller than $D$, but also the degree of smoothness of $\bff_*.$
Moreover, if $\bff_*$ has a low-dimensional composite structure considered as in \cite{horowitz2007rate}, \cite{juditsky2009nonparametric}, the convergence rate becomes faster.
For supervised learning, many studies have shown that DNN can avoid curse of dimensionality when the true regression function has a low-dimensional composite structure (\citealp{schmidt2020nonparametric, bauer2019deep, kohler2021rate}) or the support of input variables or covariates concentrate on a low-dimensional manifold (\citealp{chen2019efficient, chen2019nonparametric, schmidt2019deep, nakada2020adaptive}).
Our results  are among the first that have demonstrated  that these fine properties of DNN for supervised learning are also valid for unsupervised learning, which is an important advantage of using deep generative models compared to the ones that estimate $Q_*$ or $P_*$ directly.

The remainder of this paper is organized as follows.
In Section \ref{ssec:review}, we review recently developed theoretical results for GAN.
Section \ref{sec:model} introduces a deep generative model.
Our main results concerning the convergence rate of a sieve MLE and data perturbation are given in Section \ref{sec:asymp}.
Section \ref{sec:dist} considers a class of true distributions that can be represented as a true generator.
Experimental results and concluding remarks follow in Sections \ref{sec:exp} and \ref{sec:conclusion}, respectively.

\subsection{Related Work} \label{ssec:review}

Most works for statistical properties for deep generative models focus on GAN type estimators, which are briefly reviewed in this subsection.
In a GAN framework, \cite{arora2017generalization} firstly considered a neural network distance, a special case of IPMs, to measure the discrepancy of an estimator from the true distribution.
They noticed that a neural network distance might be so weak that GAN may not consistently estimate the true distribution.
Further studies have been conducted by \cite{zhang2018discrimination} and \cite{bai2019approximability}, who provide sufficient conditions for a neural network distance to induce the same topology as the Wasserstein metric and KL divergence.
In particular, \cite{zhang2018discrimination} obtained convergence rates of GAN estimators with respect to the bounded Lipschitz metric, which however seem to be much slower than the optimal rate.
A similar, but slightly different approach in studying a neural network distance is given in \cite{liu2017approximation}.
This work employs topological properties of neural network distances, hence important structural assumptions such as the smoothness of densities were not considered.
\cite{biau2020some} studied asymptotic properties of the original GAN developed by \cite{goodfellow2014generative}.
Rather than considering a neural network distance, they investigated how the approximation of the discriminator can affect the estimation performance with respect to the Jensen--Shannon divergence.
However, their analysis is based on the parametric assumption, that is, the number of network parameters is fixed as the sample size tends to infinity.

There is a different line of works that study asymptotic properties of GAN from a nonparametric density estimation  point of view.
For densities in a Sobolev space, \cite{singh2018nonparametric, liang2021well} derived minimax convergence rates with respect to the Sobolev IPMs which include metrics used in Sobolev (\citealp{mroueh2017sobolev}), MMD (maximum mean discrepancy; \citealp{li2017mmd}) and Wasserstein (\citealp{arjovsky2017wasserstein}) GANs.
These results are generalized in \cite{uppal2019nonparametric} using Besov IPMs.
We would also like to mention \cite{chen2020statistical}, who derived convergence rates with respect to the \Holder IPMs.
Although their convergence rate is strictly slower than the minimax rate in \cite{uppal2019nonparametric}, their results are directly applicable to GANs whose generator and discriminator network architectures are explicitly given.
However, all these works are limited to the classical paradigm where  the true distribution  possesses a smooth Lebesgue density $p_*$ and the convergence rate depends on the data dimension $D$, suffering from the curse of dimensionality.

There are some recent articles considering the convergence rate of GAN beyond the density estimation framework.
To the best of our knowledge, the set-up given in \cite{luise2020generalization} is the closest to ours.
In particular, they assumed that there exists a true generator as in our paper and there is no noise, that is, $P_* = Q_* = Q_{\bff_*}$ for some smooth function $\bff_*$.
Under this set-up they obtained a convergence rate of GAN for estimating $Q_*$ with respect to the Sinkhorn divergence (\citealp{feydy2019interpolating}).
Note that although the Sinkhorn divergence metrizes the weak convergence, it is not a standard metric  for evaluating the performance of distribution estimation and not comparable with the Wasserstein distance considered in our paper.
In particular, their convergence rate directly depends on the regularization parameter defining the Sinkhorn divergence ($\epsilon$ in their notation), which makes it unclear how tight their convergence rate is.
Furthermore, their theory does not incorporate deep neural network structures, hence cannot explain the benefit of deep generative models which adapt to various structures such as the composite one.
Also, the theory holds only when the smoothness of the true generator exceeds a certain threshold proportional to $d$.
For these reasons, the theory in \cite{luise2020generalization} has certain limitations.

\cite{schreuder2021statistical} obtained convergence rates of GAN-based estimators under the assumption that the data-generating distribution is the convolution of $Q_* = Q_{\bff_*}$ and a general noise distribution, where $\bff_* : \bbR^d \to \bbR^D$ is a smooth function; hence the data are concentrated around a small neighborhood of a manifold whose dimension is at most $d$.
Rather than assuming the existence of a true generator, \cite{huang2021error} assumed that the support of $P_*$ is a certain low-dimensional set in $\bbR^D$ and studied the convergence rate of GAN.
In both papers, the convergence rates depend on the intrinsic dimension of the true distribution rather than on the dimension $D$ of the observations.
The proofs in these papers rely on the adaptive property of the empirical measure to specific low-dimensional structures, studied in \cite{weed2019sharp} and \cite{schreuder2021bounding}. It should be noted that the intrinsic dimension considered in our paper can be smaller compared to the dimensions considered in \cite{schreuder2021statistical} and \cite{huang2021error}.

The analysis of the vanilla GAN in \cite{biau2020some} has been extended to the Wasserstein GAN in \cite{biau2021some}.
In particular, they considered DNN architectures for both the generator and discriminator classes and proved that the corresponding WGAN estimator can be arbitrarily close to the true distribution in Wasserstein distance with high probability; see Theorem 21 therein.
However, their results do not provide specific convergence rate and do not incorporate approximation error of the generator class for specific distribution families.

Finally, we would also like to mention the work by \cite{tang2022minimax} who considered the minimax convergence rate for nonparametric distribution estimation under the manifold assumption.
Although the structural assumption considered in \cite{tang2022minimax} is different from ours, they derived the minimax convergence rate for estimating a distribution supported on a submanifold of $\bbR^D$ with smooth density with respect to the Hausdorff measure.
In particular, they used a mixture of GAN estimators to achieve the minimax convergence rate.
However, it should be emphasized that GAN-based estimators considered in this subsection, including the one in \cite{tang2022minimax}, is computationally much more intractable than sieve MLEs considered in the present paper.

\subsection{Notations and Definitions}

For two real numbers $a$ and $b$, let $a\wedge b$ and $a \vee b$ be the minimum and maximum of $a$ and $b$, respectively.
$[a]$ is the largest integer less than or equal to $a$.
The inequality $a \lesssim b$ means that $a$ is less than $b$ up to a constant multiplication.
Also, denote $a \asymp b$ if $a \lesssim b$ and $b \lesssim a$.
For a vector $\bx$, the $\ell^p$-norm, $1 \leq p \leq \infty$, and the number of nonzero elements are represented as $|\bx|_p$ and $|\bx|_0$, respectively.
Let $\cB_\epsilon(\bx)$ be the Euclidean open ball of radius $\epsilon$ centered at $\bx$.
For a vector-valued function $\bff$, let $|\bff|_p$ be the map $\bx \mapsto |\bff(\bx)|_p$.
The $L^p$-norm of a function is denoted $\|\cdot\|_p$, where the domain of a function and dominating measure will be clear in the context.
The equality $c = c(A_1, \ldots, A_k)$ means that $c$ depends only on $A_1, \ldots, A_k$.
The uppercase letters, such as $P$ and $\hat P$, refer to the probability measures corresponding to the densities denoted by the lowercase letters $p$ and $\hat p$, respectively, and vice versa.
A positive real-valued function $f$ is said to be bounded from above and below if there exist positive constants $c_1$ and $c_2$ such that $c_1 \leq f(x) \leq c_2$ for every $x$.

For two probability densities $p$ and $q$, let $d_H(p, q)$ and $K(p, q) = \int \log (p/q) dP$ be the Hellinger distance and KL divergence, respectively.
The Wasserstein distance of order $r \in [1, \infty)$ between $P$ and $Q$ is denoted $W_r(P, Q)$ (\citealp{villani2003topics}).
For a function space $\cF$, $N(\delta, \cF, d)$ and $N_{[]}(\delta, \cF, d)$ denote the covering and bracketing numbers with respect to the (pseudo)-metric $d$.
For $\beta > 0$, let $\cH^\beta_M(A)$ be the class of every $\beta$-\Holder function $f: A \to \bbR$ with $\beta$-\Holder norm bounded by $M > 0$.
Let $\cH^\beta(A) = \cup_{M > 0} \cH^\beta_M(A)$ be the class of every $\beta$-\Holder function.
If there is no confusion, we simply denote them as $\cH^\beta_M$ and $\cH^\beta$.
For a vector-valued function, $\bff \in \cH^\beta$ refers that each component of $\bff$ belongs to $\cH^\beta$.
We refer to \cite{gine2016mathematical, van1996weak} for details about these definitions.

\section{Deep Generative Models}\label{sec:model}

In this section, we formally define the model $\bX = \bff(\bZ) + \bepsilon$ using a DNN.
Let $\cZ$ be an open subset of $\bbR^d$ and $\bx \mapsto \phi_{\sigma,d}(\bx)$ be the density of $d$-fold product measure of the univariate normal distribution $\cN(0, \sigma^2)$.
We often denote $\phi_{\sigma, d}$ as $\phi_\sigma$ if there is no confusion.
Let $P_{\bff, \sigma}$ be the distribution of $\bff(\bZ) + \bepsilon$, where $\bZ$ and $\bepsilon$ are independent random vectors distributed as $P_Z$ and $\cN(\bzero_D, \sigma^2 \Id_D)$, respectively.
Standard uniform or Gaussian distribution is a common choice for $P_Z$, and some general sub-Gaussian distributions are considered in \cite{luise2020generalization}.
For a class $\cF$ of functions from $\cZ$ to $\bbR^D$ and two positive numbers $\sigma_{\min} < \sigma_{\max}$, we consider a class of probability distributions
\be\label{eq:model}
	\cP = \Big\{ P_{\bff, \sigma}: \bff \in \cF, \sigma \in [\sigma_{\min}, \sigma_{\max}] \Big\}.
\ee
Recall that $Q_\bff$ is the distribution of $\bff(\bZ)$, which is often called the pushforward measure of $P_Z$ by the map $\bff:\cZ \to \bbR^D$.
If $\sigma > 0$, $P_{\bff, \sigma}$ has the Lebesgue density
\be \label{eq:model-pdf}
	p_{\bff, \sigma}(\bx) = \int \phi_\sigma \big(\bx - \bff (\bz)\big) dP_Z(\bz) = \int \phi_\sigma(\bx - \bu) dQ_\bff(\bu).
\ee

The function class $\cF$ is modeled via a DNN.
We adopt the definitions and notations in \cite{schmidt2020nonparametric}.
Let $\rho(x) = x \vee 0$ be the ReLU activation function.
For a vector $\bv = (v_1, \ldots, v_r)^T \in \bbR^r$, define $\rho_\bv: \bbR^r \to \bbR^r$ as $\rho_\bv(\bz) = (\rho(z_1 - v_1), \ldots, \rho(z_r - v_r))^T$ for $\bz = (z_1, \ldots, z_r)^T$.
A neural network with network architecture $(L, \bp)$ is any function of the form
\be\label{eq:dnn}
	\bff: \bbR^{p_0} \to \bbR^{p_{L+1}}, 
	\quad \bz \mapsto \bff(\bz) = W_L \rho_{\bv_L} W_{L-1} \rho_{\bv_{L-1}} \cdots W_1 \rho_{\bv_1} W_0 \bz,
\ee
where $W_i \in \bbR^{p_{i+1} \times p_i}$, $\bv_i \in \bbR^{p_i}$ and $\bp = (p_0, \ldots, p_{L+1}) \in \bbN^{L+2}$.
We will consider model \eqref{eq:model} with the class $\cF = \cF(L, \bp, s, K)$, where $\cF(L, \bp, s, K)$ is the collection $\bff$ of the form \eqref{eq:dnn} satisfying
\bean
	\max_{j=0, \ldots,L} |W_j|_\infty \vee |\bv_j|_\infty \leq 1, \quad
	\sum_{j=1}^L |W_j|_0 + |\bv_j|_0 \leq s, \quad
	\| |\bff|_\infty \|_\infty \leq K,
\eean
$p_0 = d$ and $p_{L+1}=D$.
Here, $|W_j|_\infty$ and $|W_j|_0$ denote the maximum-entry norm and the number of nonzero elements of the matrix $W_j$, respectively.

The statements of main theorems and corollaries in Section \ref{sec:asymp} are non-asymptotic; they hold for any fixed $n\geq 1$.
However, it would be convenient to regard quantities $(\sigma_{\min}, L, \bp, s)$ as sequences depending on the sample size $n$, while $(\sigma_{\max}, K)$ remain as fixed constants.
In this sense, it would be precise to denote $(\sigma_{\min}, L, \bp, s)$ and $(\cF, \cP)$ as $(\sigma_{\min, n}, L_n, \bp_n, s_n)$ and $(\cF_n, \cP_n)$, respectively.
For simplicity, we suppress the subscript when the dependency on $n$ is obvious contextually.
Throughout this paper, the model \eqref{eq:model} with $\cF = \cF(L, \bp, s, K)$ will be called a \textit{deep generative model} with ReLU activation function.

From another viewpoint, the density of the form \eqref{eq:model-pdf} is a mixture of normal distributions.
Note that mixtures of normal densities are frequently used in nonparametric statistics to model smooth densities.
In particular, an arbitrary smooth density can be approximated by normal mixtures as shown in \cite{ghosal2007posterior, shen2013adaptive}.
Based on this, it can be shown that a Bayes estimator with a Dirichlet process prior and a sieve MLE achieve the minimax optimal convergence rate up to a logarithmic factor when the true density belongs to a \Holder class.
However, the model complexity of normal mixtures required to approximate an arbitrary smooth density, often expressed through the metric entropy, grows rapidly as the dimension $D$ increases which results in slow convergence rates.
This large complexity is mainly because the mixing distribution can be of any form.
Hence, such a large class of normal mixtures might not be useful for analyzing high-dimensional data.
Note that model \eqref{eq:model} is parametrized by the generator $\bff$ rather than a mixing distribution.
Consequently, the complexity of the model \eqref{eq:model} can be expressed through the metric entropy of the generator class $\cF$, which is detailed in Lemma \ref{lem:bracket-entropy}.

\section{Convergence Rate of a Sieve MLE}
\label{sec:asymp}

Our main theoretical results are given in this section.
We first present assumptions on the data-generating distribution $P_*$.
Then, we derive the convergence rate of a sieve MLE for $p_*$ with respect to the Hellinger distance in the deep generative model.
We next obtain the convergence rate of the corresponding sieve MLE of $Q_*$ under the Wasserstein distance. 
Our strategy of deriving the convergence rate is as follows.
We first  derive a convergence rate of a sieve MLE $\hat{p}$ of $p_*$, the Lebesgue density of $P_*,$  and then recover the corresponding convergence rate of $\hat Q$ to $Q_*$. 
However, this strategy only works when $\sigma_*$ is not too small.
If $\sigma_*$  is very small,  technical difficulty arises because the density $p_*$ peaks around a small neighborhood of $\bff_*(\cZ)$, the likelihood therefore becomes picky and unstable, and a sieve MLE is expected to behave badly. 
For this case, we propose a novel data perturbation technique to derive the convergence rates for $Q_*$ under this small $\sigma_*$ regimes.

As mentioned earlier, our main theorems are non-asymptotic in the sense that they hold for any fixed $n \geq 1$.
More specifically, Theorem \ref{thm:W-rate} is stated with the form of 
\be \label{eq:main-statement}
	P_*\big( W_1 (\hat Q, Q_*) > \epsilon_n \big) \leq \delta_n, \quad n \geq 1
\ee
for some sequences $\delta_n$ and $\epsilon_n$ with $\epsilon_n \ll \delta_n$.
The interpretation of this statement is clear: for any fixed $n\geq 1$, once $\epsilon_n$ and $\delta_n$ are small enough, the Wasserstein distance between $\hat Q$ and $Q_*$ will be small with high probability.
Furthermore, since $\hat Q$ and $Q_*$ are supported on a bounded set, the probabilistic statement \eqref{eq:main-statement} implies that $\E W_1(\hat Q, Q_*) \lesssim \epsilon_n \vee \delta_n$ for every $n \geq 1$. 
Similar interpretations also hold for assumptions of the Theorems on the noise $\sigma_*$, that is, for every sample size, there is a sufficient condition on the noise $\sigma_*$ for which the probabilistic bound \eqref{eq:main-statement} holds. 
Given the non-asymptotic nature of our results, the true data-generating distribution can be interpreted in similar fashions. 
For any given sample size $n\geq 1$, the true data-generating distribution is given by a true $P_*$ induced from the true generator $\bff_*$ and  some true noise level $\sigma_* \in [\sigma_{\min}, \sigma_{\max}] $ with some appropriate assumptions on $\sigma_{\min}$ and $\sigma_{\max}$. 
The assumptions on  $\sigma_{\min}$ and $\sigma_{\max}$ may vary with the sample size $n$. 

Note that such non-asymptotic statements and interpretation can be frequently found in modern statistical theory. 
For example, in a high-dimensional linear regression set-up, the assumption on the dimension and/or the magnitude of the regression coefficients $\beta_*$ may change with the sample size (\citealp{buhlmann2011statistics, wainwright2019high}). 
When the sample size is large, for example, the absolute value of the first component of $\beta_*$ may be assumed to be large.
For any fixed $n\geq 1$, however, there is one true data-generating distribution with the true parameter satisfying the appropriate assumption.  
In this set-up, many statistical theories take the form $P_* (\|\hat\beta - \beta_* \| > \epsilon_n ) \leq \delta_n$, which is quite similar to \eqref{eq:main-statement}.

\subsection{Assumption on the True Distribution}

Since we consider a deep generative model \eqref{eq:model}, it is natural to assume that $P_* = P_{\bff_*, \sigma_*}$ for some true generator $\bff_*$ and $\sigma_* \geq 0$, or more precisely, $P_*$ is the convolution of $Q_* = Q_{\bff_*}$ and $\cN(\bzero_D, \sigma_*^2 \Id_D)$.
In particular, we assume that $\bff_*$ is a structured function that can be efficiently approximated by DNN functions (\citealp{yarotsky2017error, telgarsky2016benefits, petersen2018optimal, ohn2019smooth, imaizumi2019deep, nakada2020adaptive}).
For example, $\bff_*$ can belong to a certain class $\cF$ of smooth composite functions.
In Section \ref{sec:dist}, we will show that the corresponding distribution class $\{ Q_\bff: \bff \in \cF\}$ is large enough to include the classical class of nonparametric smooth densities and densities supported on a lower-dimensional smooth manifolds as special cases.

Note that the generator $\bff_*$ is not identifiable in general.
For example, even for a linear factor model where $\bff_*(\bZ)= \bA\bZ$ for a $D\times d$ matrix $\bA$, $\widetilde{\bff}(\bZ)=-\bA\bZ$ has the same distribution as $\bff_*(\bZ).$ 
However, the mixing distribution $Q_*$ is identifiable under mild assumptions, \eg\ \cite{bruni1985identifiability}.

\subsection{A Sieve MLE}

Since the parameter space specifying the model \eqref{eq:model} depends on the sample size $n$, the model can be regarded as a sieve approximating the true distribution.
Then, an estimator can be obtained via a maximum likelihood principle.
The corresponding estimator is often called a sieve MLE (\citealp{geman1982nonparametric}).
To be specific, let $\ell_n(\bff, \sigma) = \sum_{i=1}^n \log p_{\bff, \sigma}(\bX_i)$ be the log-likelihood function.
For a given sequence $\eta_n \downarrow 0$, a sieve MLE is any estimator $(\hat \bff, \hat \sigma) \in \cF \times [\sigma_{\min}, \sigma_{\max}]$ satisfying
\be \label{eq:sieveMLE}
	\ell_n(\hat \bff, \hat \sigma) \geq \sup_{\bff \in \cF, \sigma \in [\sigma_{\min}, \sigma_{\max}]} \ell_n(\bff, \sigma) - \eta_n
\ee
and let $\hat p = p_{\hat\bff, \hat\sigma}$.
We do not abbreviate the subscript $n$ for the rate sequence such as $\eta_n$ and $\epsilon_n$.
The sequence $\eta_n$ allows that strict maximization, which is infeasible in most applications of deep learning,
is not necessary.
It would be more desirable to consider an estimator which is obtained by a specific algorithm such as the gradient decent method.
Unfortunately, it is challenging to study statistical properties of an algorithm-specific estimator in deep learning.
To the best of our knowledge, the convergence rate of an algorithm-specific estimator have not been studied in deep learning contexts.
We also do not consider algorithmic issues in this paper, and assume that a sieve MLE satisfying \eqref{eq:sieveMLE} is available.
There are various computational algorithms targeting a sieve MLE in deep generative models, \eg\ \cite{burda2016importance, kim2020casting}.

\subsection{Hellinger Convergence Rate of a Sieve MLE of $p_*$}

Under general conditions, convergence rates of sieve MLEs with respect to the Hellinger metric are well established in \cite{wong1995probability}.
The key technique to derive convergence rates is to bound the Hellinger bracketing number of the density space for which many techniques are known for various classes of regular functions, see \cite{van1996weak}.
Roughly, the convergence rate $\epsilon_n$ can be achieved if $\log N_{[]}(\delta, \cP, d_H) \lesssim n\epsilon_n^2$.
Metric entropies of deep neural networks are also well-known in recent articles, see Lemma 5 of \cite{schmidt2020nonparametric}.
The following lemma provides a relation between the Hellinger bracketing number of $\cP$ and the metric entropy of $\cF,$
which plays a crucial role in deriving the convergence rate of a sieve MLE $\hat p.$
Below, we do not try to optimize constants which are not essential for deriving convergence rates.

\begin{lemma}\label{lem:bracket-entropy}
Let $\cF$ be a class of functions from $\cZ$ to $\bbR^D$ such that $\| |\bff|_\infty \|_\infty \leq K$ for every $\bff \in \cF$.
Let $\cP = \{ P_{\bff, \sigma}: \bff \in \cF, \sigma \in [\sigma_{\min}, \sigma_{\max}]\}$ with $\sigma_{\min} \leq 1$.
Then, there exist constants $c=c(D, K, \sigma_{\max}), c'=c'(D, K, \sigma_{\max})$ and $\delta_*=\delta_*(D)$ such that
\be \label{eq:bracket}
	\log N_{[]}(\delta, \cP, d_H)
	\leq \log N\left( c \sigma_{\min}^{D+3} \delta^4, \cF, \| |\cdot|_\infty \|_\infty \right)
	+ \log\left(\frac{c'}{\sigma_{\min}^{D+2} \delta^4}\right)
\ee
for every $\delta \in (0, \delta_*]$.
\end{lemma}
\medskip

\begin{remark}
Note that for a class of general normal location mixtures $\int \phi_\sigma(\bx - \bz) dP(\bz)$ parametrized by the mixing distribution $P$ and scale parameter $\sigma$, the bracketing entropy scales as a polynomial order in $\sigma^{-1}$ as $\sigma \to 0$.
Specifically, Corollary B1 of \cite{shen2013adaptive} gives an upper bound for the $\delta$-bracketing entropy of the class $\{\bx \mapsto \int \phi_\sigma(\bx - \bz) dP(\bz): P([-K, K]^D)=1\}$, which is at least of order $O((\sigma^{-1} \vee \log \delta^{-1})^D)$.
This bound would give a nearly parametric convergence rate of a sieve MLE provided that the model is well-specified and $\sigma_{\min}$ is bounded away from zero.
However, the entropy bound of \cite{shen2013adaptive} grows rapidly as $\sigma_{\min} \to 0,$
which is problematic since we are interested in the case that $\sigma_{\min}$ converges to 0.
In contrast, the right hand side of \eqref{eq:bracket} depends on $\sigma_{\min}$ only through a logarithmic function. 
Hence, the entropy bound \eqref{eq:bracket} is much smaller than that of \cite{shen2013adaptive}
when $\sigma_{\min}$ is small, 
provided that $N(\delta, \cF, \| |\cdot|_\infty \|_\infty)$ is of a polynomial order in $\delta$.
If $\cF = \cF(L, \bp, s, \infty)$ with $\|\bp\|_\infty = O(n^a)$ for some constant $a>0$ and $L = O(\log n)$, for example, $\log N (\delta, \cF, \| |\cdot|_\infty \|_\infty)$ is bounded by a multiple of $s \{(\log n)^2 + \log \delta^{-1}\}$, as shown in Lemma 5 of \cite{schmidt2020nonparametric}. Consequently, $\log N_{[]}(\delta, \cP, d_H)$ is of order $s\{(\log n)^2 + \log \delta^{-1} + \log \sigma_{\min}^{-1} \}.$
\end{remark}

Utilizing Lemma \ref{lem:bracket-entropy}, the next theorem provides convergence rates of a sieve MLE of $p_*$ with respect to the Hellinger metric in terms of the entropy bound and approximation error $\delta_{\rm app}$ of the sieve $\cF$.

\begin{theorem} \label{thm:rate-general}
Let $\cF, \cP$ and $\delta_*=\delta_*(D)$ be given as in Lemma \ref{lem:bracket-entropy}, and $n \geq 1$.
Suppose that $\log N(\delta, \cF, \| |\cdot|_\infty \|_\infty) \leq s\{ A + 1\vee\log \delta^{-1}\}$ for every $\delta > 0$.
Assume also that there exists $\bff \in \cF$ such that $\| |\bff - \bff_*|_\infty \|_\infty \leq \delta_{\rm app}$.
Furthermore, suppose that $s\geq 1$, $A\geq 1$, $\sigma_{\min} \leq 1$, $\delta_{\rm app} \leq 1$ and $\sigma_* \in [\sigma_{\min}, \sigma_{\max}]$.
Then, a sieve MLE $\hat p$ defined through \eqref{eq:sieveMLE} satisfies that 
\be\label{eq:rate-general}
	P_* \Big( d_H(\hat p, p_*) > \epsilon_n^* \Big) 
	\leq 5 e^{-C_1 n \epsilon_n^{*2}} + \frac{C_2}{n}
\ee
provided that $\eta_n \leq \epsilon_n^{*2}/6$ and $\epsilon_n^* \leq \sqrt{2} \delta_*$, where
\bean
	\epsilon_n^* = C_3 \left(\sqrt{\frac{s \{A + \log (n/\sigma_{\min})\}}{n}} 
	\vee \frac{\delta_{\rm app}}{\sigma_*}\right),
\eean
$C_1$ is an absolute constant, $C_2 = C_2(D)$ and $C_3=C_3(D, K, \sigma_{\max})$.
\end{theorem}

Using Theorem \ref{thm:rate-general}, we can derive the convergence rate of a sieve MLE
of deep generative models for various $\bff_*.$ As an illustrative example,
suppose that $\bff_* \in \cH^\beta_K\big((0,1)^d\big)$ for some positive constants $\beta$ and $K$.
Since a smooth function can be efficiently approximated by DNN, one can obtain a convergence rate as in the following corollary.
We omit the proof because it is a special case of Corollary \ref{cor:composition} with $q=0$ and $d=d_0=t_0.$

\begin{corollary} \label{cor:rate-smooth}
Suppose that $\bff_* \in \cH^\beta_K\big((0,1)^d \big)$, $\sigma_* = n^{-\alpha}$ and $\sigma_{\min} = n^{-\gamma}$ for some $\beta, K > 0$ and $0 \leq \alpha \leq \gamma$.
Then, there exists a network architecture $\cF = \cF(L, \bp, s, K)$ (depending only on $(n, d, \beta, K)$) such that a sieve MLE $\hat p$ satisfies
\bean
	P_* \Big( d_H(\hat p, p_*) > \epsilon_n^* \Big) \leq 5 e^{-C_1 n \epsilon_n^{*2}} + \frac{C_2}{n}
\eean
provided that $\eta_n \leq \epsilon_n^{*2} /6$ and $\epsilon_n^* \leq \sqrt{2} \delta_*$, where $C_1$, $C_2=C_2(D), \delta_*=\delta_*(D)$ are constants in Theorem \ref{thm:rate-general} and $\epsilon_n^* = C n^{-(\beta - d\alpha) / (2\beta + d)} (\log n)^{3/2}$ with $C=C(\alpha, \beta, \gamma, d, D, K, \sigma_{\max})$.
\end{corollary}

The statement of Corollary \ref{cor:rate-smooth} is overly simplified to illustrate the role of the dimension, smoothness and noise level in the convergence rate.
In particular, the rate gets faster as the noise level increases.
This seemingly paradoxical phenomenon  occurs because $p_*$ gets smoother as $\sigma_*$ increases.
On the other hand, for a very small value of $\sigma_*$,  for  consistent estimation of $p_*$ it is necessary to have very accurate approximation of  $\bff_*$. 
For this purpose, it is inevitable to increase the number of nonzero network parameters, which leads to an increase in the estimation error.
In the set-up of Corollary \ref{cor:rate-smooth}, the number of nonzero network parameters $s$  needed for a suitable degree of approximation is of order $n^{d(2\alpha+1) / (2\beta+d)}$ up to a logarithmic factor.
Note that the condition $\beta > d\alpha$ is equivalent to that $d(2\alpha+1)/(2\beta+d)$ is strictly smaller than 1.
That is, when $\beta \leq d \alpha$, too many nonzero coefficients are needed to ensure that the approximation error is sufficiently small.
Consequently, Theorem \ref{thm:rate-general} does not even guarantee the consistency.
The case for a very small $\sigma_*$ will be handled in Section \ref{ssec:perturbation} with a novel data perturbation technique.
Before that, we assume that $\sigma_*$ is not too small.

When $\bff_*$ has a low-dimensional structure,
the convergence rate in Corollary \ref{cor:rate-smooth} can be significantly improved. 
We consider the composition structure with low-dimensional smooth component functions as described in Section 3 of \cite{schmidt2020nonparametric}.
Specifically, we consider a function $\bff$ of the form
\be \label{eq:composition}
	\bff = \bg_q \circ \bg_{q-1} \circ \cdots \circ \bg_1 \circ \bg_0
\ee
with $\bg_i: (a_i, b_i)^{d_i} \to (a_{i+1}, b_{i+1})^{d_{i+1}}$.
Here, $d_0=d$ and $d_{q+1}=D$.
Denote by $\bg_i = (g_{i1}, \ldots, g_{id_{i+1}})^T$ the components of $\bg_i$ and let $t_i$ be the maximal number of variables on which each of the $g_{ij}$ depends.
Let $\cG(q, \bd, \bt, \bbeta, K)$ be the collection of functions of the form \eqref{eq:composition} satisfying $g_{ij} \in \cH^{\beta_i}_K\big( (a_i, b_i)^{t_i} \big)$ and $|a_i| \vee |b_i| \leq K$, where $\bd = (d_0, \ldots, d_{q+1})^T$, $\bt = (t_0, \ldots, t_q)^T$ and $\bbeta = (\beta_0, \ldots, \beta_q)^T$.
It would be convenient to regard quantities $(q, \bd, \bt, \bbeta, K)$ as constants.
Let
\bean
	\widetilde \beta_j = \beta_j \prod_{l=j+1}^q (\beta_l \wedge 1), 
	\quad
	j_* = \argmax_{j \in \{0, \ldots,q\}} \frac{t_j}{\widetilde \beta_j},
	\quad
	\beta_* = \widetilde\beta_{j_*},
	\quad
	t_* = t_{j_*}.
\eean
We call $t_*$ and $\beta_*$ as the \textit{intrinsic dimension} and \textit{smoothness} of $\bff$ (or of the function class $\cG(q, \bd, \bt, \bbeta, K))$, respectively.

Any function $\bff$ in $\cG(q, \bd, \bt, \bbeta, K)$ can be efficiently approximated by a DNN as detailed in Lemma \ref{lem:composition-approx}.
The proof can be easily deduced from the proof of Theorem 1 in \cite{schmidt2020nonparametric}.
Then, Corollary \ref{cor:composition} provides the convergence rates of $\hat{p}$ when $\bff_*$ has the composition structure.

\begin{lemma}\label{lem:composition-approx}
Suppose that $\bff_* \in \cG(q, \bd, \bt, \bbeta, K)$.
Then, for every $\delta \in (0,1)$, there exists a network $\cF = \cF(L, \bp, s, K\vee 1)$ with $L \leq c_1 \log \delta^{-1}$, $|\bp|_\infty \leq c_2 \delta^{-t_* / \beta_*}$, $s \leq c_3 \delta^{-t_* / \beta_*} \log \delta^{-1}$ satisfying $\| |\bff - \bff_* |_\infty \|_\infty \leq \delta$ for some $\bff \in \cF$, where $c_j = c_j(q, \bd, \bt, \bbeta, K)$ for $j\in\{1,2,3\}$.
\end{lemma}

\begin{corollary} \label{cor:composition}
Suppose that $\bff_* \in \cG(q, \bd, \bt, \bbeta, K)$, $\sigma_* \in [\sigma_{\min}, \sigma_{\max}]$, $\sigma_{\min} \leq 1$ and 
\bean
	\delta_{\rm app} \defeq \left(\frac{\sigma_*^2}{n}\right)^{\frac{\beta_*}{2\beta_* + t_*}} \leq 1.
\eean
Let $\cF = \cF(L, \bp, s, K\vee 1)$ with $L = [c_1 \log \delta_{\rm app}^{-1}]$, $p_0= \cdots =p_{L+1} = [c_2 \delta_{\rm app}^{-t_* / \beta_*}]$, $s = [c_3 \delta_{\rm app}^{-t_* / \beta_*} \log \delta_{\rm app}^{-1}] + 1$, where $c_j=c_j(q, \bd, \bt, \bbeta, K)$, $j\leq 3$, are constants in Lemma \ref{lem:composition-approx}.
Define $\delta_*=\delta_*(D)$ and $\epsilon_n^*$ as in Theorem \ref{thm:rate-general} with $A = c_4 (\log n)^2$, where $c_4 = c_4(q, \bd, \bt, \bbeta, K)$ as specified in the proof.
If $\eta_n \leq \epsilon_n^{*2}/6$ and $\epsilon_n^* \leq \sqrt{2} \delta_*$, a sieve MLE $\hat p$ satisfies \eqref{eq:rate-general}.
\end{corollary}

In Corollary \ref{cor:composition}, the approximation error $\delta_{\rm app}$ is chosen so that
\bean
	\epsilon_n^* \asymp \sqrt{\frac{s}{n}} \asymp \frac{\delta_{\rm app}}{\sigma_*}
\eean
up to a logarithmic factor. 
More precisely, if $\sigma_* = n^{-\alpha}$ and $\sigma_{\min} = n^{-\gamma}$ for some $\gamma \geq \alpha \geq 0$, we have 
\bean
	\epsilon_n^* = C n^{-\frac{\beta_* - t_* \alpha}{2\beta_* + t_*}} (\log n)^{3/2},
\eean
where $C = C(q, \bd, \bt, \bbeta, K, D, \sigma_{\max}, \alpha, \gamma)$.
As one can see,  the dimension $d$ in the convergence rate of Corollary \ref{cor:rate-smooth} is replaced by the intrinsic dimension $t_*$.
If $t_*$ is much smaller than $d$, the improvement from the structural assumption would be significant.

\subsection{Wasserstein Convergence Rate of a Sieve MLE of $Q_*$}

Since we are primarily interested in estimating $Q_* = Q_{\bff_*}$, in this section we consider the problem of estimating $Q_*$ and  utilize the $L^1$-Wasserstein metric as an evaluation metric. Given a sieve MLE \eqref{eq:sieveMLE}, an estimator can be easily constructed as $\hat Q = Q_{\hat \bff}$.
Note that obtaining an upper bound of $W_1(\hat Q, Q_*)$ from $d_H(p_*, \hat p)$ is a kind of deconvolution problem.
A sharp bound for this problem is established in Section 2.3 of \cite{nguyen2013convergence} when $\sigma_*$ and $\hat\sigma$ are bounded away from zero.
For example, with the $L^2$-Wasserstein metric, a sharp bound $W_2^2(Q_*, \hat Q) \lesssim \{ -\log d_H(p_*, \hat p) \}^{-1}$ is achievable, see Theorem 2 of \cite{nguyen2013convergence}.
Hence, even when $d_H(p_*, \hat p)$ decays with a polynomial rate, one can only expect a very slow convergence rate for $W_2(Q_*, \hat Q)$; see also \cite{fan1991optimal} and \cite{alexander2009deconvolution} for a more formal statistical theory for the deconvolution.
Such a logarithmic minimax rate can also be found in a slightly different but closely related problem.
More specifically, \cite{genovese2012manifold} considered the problem of estimating the support of the singular distribution $Q_*$ and obtained a lower bound $(\log n)^{-1}$ for the minimax optimal rate under the Hausdorff distance, see Theorem 8 therein.
The slow minimax rates in the deconvolution and manifold estimation problems are closely related to the super-smoothness of the normal density.
Here, a super-smoothness density roughly means that the tail of the Fourier transform of the density decays faster than any inverse polynomial, see Theorem 2 of \cite{nguyen2013convergence}.
For a small value of $\sigma_*$, however, a much faster convergence rate is achievable because $\phi_{\sigma_*}$ is no longer smooth.

Before studying the convergence rate, it would be worth addressing the identifiability issue.
Since $p_*(\bx) = \int \phi_\sigma(\bx-\bu) dQ_*(\bu)$, $Q_*$ can be understood as a mixing distribution for the data distribution $P_*$ with the normal kernel.
In this case, $Q_*$ is identifiable under very mild conditions, see \cite{bruni1985identifiability}.
However, the identifiability does not guarantee an efficient estimation of $Q_*$.
In some identifiable mixture models, the minimax convergence rate for estimating the mixing distribution can be very slow, see \cite{wei2022convergence}.
A stronger identifiability condition is often necessary for obtaining a fast convergence rate of the mixing distribution.

In this subsection, we impose a strong identifiability condition through the reach of a manifold, which is introduced by \cite{federer1959curvature} and frequently used in manifold estimation contexts. 
For a set $\cM \subset \bbR^D$ and $r>0$, let $\cM^r = \cM \oplus \cB_r(\bzero_D)$ be the $r$-enlargement of $\cM$, where $\oplus$ stands for the Minkowski sum.
The reach of a closed set $\cM$, denoted as reach$(\cM)$, is defined as the supremum of $r$ with the property that any point in $\cM^r$ has a unique Euclidean projection onto $\cM$.

In forthcoming Theorem \ref{thm:rate-W}, we assume that reach$(\cM_*)$ is bounded below by a positive number, where $\cM_*$ is the closure of $\bff_*(\cZ)$.
This is one of the most important assumption in manifold estimation literature (\citealp{aamari2019nonasymptotic, divol2021minimax, puchkin2022structure, tang2022minimax}).
Note that even consistent estimation of $Q_*$ may not be possible if reach$(\cM_*) = 0$, as shown in \cite{berenfeld2019density}.

\begin{theorem}\label{thm:rate-W}
Let $\cM_*$ be the closure of $\bff_*(\cZ)$.
Suppose that $\| |\bff_*|_\infty\|_\infty \leq K$ for a constant $K$.
Also, assume that $\cM_*$ does not have an interior point in $\bbR^D$, and reach$(\cM_*) = r_*$ for some constant $r_*>0$.
Then, $d_H(p_{\bff, \sigma}, p_*) \leq \epsilon \leq 1$ and $\| |\bff|_\infty\|_\infty \leq K$ imply that $W_1(Q_\bff, Q_*) \leq C(\epsilon + \sigma_* \sqrt{\log \epsilon^{-1}})$, where $C = C(D, K, r_*)$.
\end{theorem}

Theorem \ref{thm:rate-W} guarantees that $W_1(\hat Q, Q_*) \lesssim d_H(\hat p, p_*) + \sigma_*$ up to a logarithmic factor.
Since we have already obtained a rate for $d_H(\hat p, p_*)$, it is possible to obtain a Wasserstein convergence rate for estimating $Q_*$.
For example, when $\bff_* \in \cH^\beta_K\big((0,1)^d\big),$ Corollary \ref{cor:rate-smooth} together with Theorem \ref{thm:rate-W}
implies that there exists a sieve of deep generative models with which
the convergence rate of $W_1(\hat Q, Q_*)$ is $ O_p\big(n^{-(\beta - d\alpha) / (2\beta + d)} (\log n)^{3/2} \vee \sigma_* \sqrt{\log n}\big).$

\begin{remark}
Note that Theorem \ref{thm:rate-W} does not require $\bff_*(\cZ)$ to be a topological or smooth manifold.
For example, $\bff_*(\cZ)$ can be a union of two manifolds with different dimensions.
\end{remark}

\subsection{Data Perturbation} \label{ssec:perturbation}

When $\sigma_*$ is too small, the convergence rates of $d_H(\hat{p},p_*)$ obtained in Corollaries \ref{cor:rate-smooth} and \ref{cor:composition} do not even converge to 0 as the sample size increases:
in Corollary \ref{cor:composition}, for example,  when $\sigma_* \ll n^{-\beta_*/t_*}$,  with  $\beta_*<t_* \alpha$.
Under these regimes,  $p_*$ peaks around a small neighborhood of $\bff_*(\cZ)$ and  the singularity exacerbates, thus a sieve MLE does not behave well. 
In an extreme case where $\sigma_*=0,$ $P_*$ itself is a singular measure and likelihood approaches cannot be justified via minimizing the Kullback--Leibler (KL) divergence. 

To overcome these difficulties, we consider the perturbed observations $\widetilde\bX_i = \bX_i + \widetilde \bepsilon_i$, where $\widetilde \bepsilon_i \sim \cN(\bzero_D, \widetilde\sigma^2 \Id_D)$ is an artificial noise vector.
Note that $\widetilde \bX_1, \ldots, \widetilde\bX_n$ can be understood as \iid\ observations from the true distribution $\widetilde P_* = P_{\bff_*, \widetilde\sigma_*}$, where $\widetilde\sigma_*^2 = \sigma_*^2 + \widetilde \sigma^2$.
Let $(\hat \bff_{\rm per}, \hat\sigma_{\rm per})$ be a sieve MLE based on the perturbed observation $\widetilde\bX_1, \ldots, \widetilde\bX_n$.
Also, define $\hat P_{\rm per} = P_{\hat\bff_{\rm per}, \hat\sigma_{\rm per}}$ and $\hat Q_{\rm per} = Q_{\hat\bff_{\rm per}}$ accordingly.

Once we use $\hat Q_{\rm per}$ as an estimator for $Q_*$, we have $W_1(\hat Q_{\rm per}, Q_*) \lesssim \widetilde \epsilon_n + \widetilde \sigma_* \sqrt{\log \widetilde\epsilon_n^{-1}}$ by Theorem \ref{thm:rate-W}, where $\widetilde \epsilon_n = d_H(\hat p_{\rm per}, \widetilde p_*)$.
As $\widetilde\sigma$ increases, note that $\widetilde \epsilon_n$ decreases while $\widetilde \sigma_*$ increases.
Thus, the convergence rate for $W_1(\hat Q_{\rm per}, Q_*)$ can be optimized by choosing $\widetilde{\sigma}$ accordingly, which is summarized in the following theorem.

\begin{theorem}\label{thm:W-rate}
Let $n \geq 1$, $\bff_* \in \cG(q, \bd, \bt, \bbeta, K)$, $\sigma_* \in [\sigma_{\min}, \sigma_{\max}]$, $\sigma_* = n^{-\alpha}$ and $\sigma_{\min} = n^{-\gamma}$ for some $\gamma \geq \alpha \geq0$.
Assume that $Q_*(\cM_*)=1$ and reach$(\cM_*)\geq r_*$, where $r_* > 0$ and $\cM_*$ is the closure of $\bff_*(\cZ)$.
Then, there exists a network architecture $\cF = \cF(L, \bp, s, K)$ (depending only on $(n, q, \bd, \bt, \bbeta, K)$) such that sieve MLEs $\hat p_{\rm per}$ and $\hat Q_{\rm per}$ based on the perturbed observation $\widetilde\bX_i = \bX_i + \widetilde \bepsilon_i$, with $\widetilde\bepsilon_i \sim \cN(\bzero_D, n^{-\beta_*/(\beta_*+ t_*)} \Id_D)$, satisfies
\be\label{eq:rate-W}
	P_*\left( W_1(\hat Q_{\rm per}, Q_*) > C_3 \Big( \epsilon_n^* + \sigma_* \sqrt{-\log \epsilon_n^*} \Big) \right)
	\leq 5 e^{-C_1 n\epsilon_n^{*2}} + \frac{C_2}{n},
\ee
where
\bean
	\epsilon_n^* = \left\{ \begin{array}{ll}
		C_4 n^{-\frac{\beta_* - t_* \alpha}{2\beta_* + t_*}} (\log n)^{3/2} & ~~\text{if $\alpha < \beta_*/ \{2(\beta_* + t_*)\}$,}
		\\
		C_5 n^{-\frac{\beta_*}{2(\beta_* + t_*)}} (\log n)^{3/2} & ~~\text{otherwise},
	\end{array}\right.
\eean
 $C_1$ is an absolute constant, $C_2 = C_2(D)$, $C_3 = C_3(D, K,  r_*)$, $C_4$ and $C_5$ depend only on $(q, \bd, \bt, \bbeta, K, D, \sigma_{\max}, \alpha, \gamma)$.
\end{theorem}

To the best of our knowledge, our main result (Theorem \ref{thm:W-rate}) is the first theory considering the Wasserstein convergence of $\hat Q$ in a deep generative model with the intrinsic dimension and smoothness of $\bff_*$.
Most existing theories consider GAN type estimators and have derived convergence rates that depend on either the intrinsic dimension alone or $D$.

If $\alpha < \beta_* / \{2(\beta_* + t_*)\}$, we have $\sigma_* \gg \epsilon_n^*$ so $\sigma_* \sqrt{-\log \epsilon_n^*}$ in the left hand side of \eqref{eq:rate-W} is the dominating term.
Therefore, regardless of $\alpha < \beta_* / \{2(\beta_* + t_*)\}$, we conclude that
\be \label{eq:rate-final}
	W_1 (\hat Q_{\rm per}, Q_*)
	\lesssim n^{-\frac{\beta_*}{2(\beta_* + t_*)}} (\log n)^{3/2}
	+ \sigma_* \sqrt{\log n}
\ee
with high probability.
Since $W_1(\hat Q_{\rm per}, Q_*)$ is a bounded random variable, its expectation can also be easily bounded by a multiple of the right hand side of \eqref{eq:rate-final}.

It can be easily deduced from the proof that the data perturbation improves the convergence rate only when $\sigma_*\lesssim n^{-\beta_*/2(\beta_*+ t_*)}$.
Note that the level of perturbation and the network architecture in Theorem \ref{thm:W-rate} depend on the unknown quantities $(\beta_*, t_*, \sigma_*)$.
In other words, our results are non-adaptive to the unknown structure.
Hence, the network architectures and $\widetilde\sigma$ are tuning parameters that should be carefully chosen.
To obtain an estimator adaptive to the unknown structure, two approaches are known in the literature for the deep supervised learning.
The first one is a penalized likelihood approach such as the lasso and non-convex penalties as considered in \cite{ohn2022nonconvex}.
Alternatively, Bayesian approaches can be utilized to obtain an adaptive estimator, see \cite{polson2018posterior, ohn2021adaptive}.
Although these papers studied nonparametric regression, it would be possible to extend their approaches to deep generative models to obtain an adaptive estimator.
In practice, there are several heuristic methods to select network architectures (\citealp{salimans2016improved, arjovsky2017wasserstein, radford2016unsupervised}).
The variance $\widetilde \sigma^2$ of the additional noise is 1-dimensional, hence it can also be tuned based on the validation error without much difficulty; see Section \ref{sec:exp} for details.

After the original version of this article was drafted, the first author investigated the lower bound for the minimax optimal convergence rate with the structural assumption considered in Theorem \ref{thm:W-rate}, which is now available in \cite{chae2022rates}.
Specifically, he obtained a lower bound $n^{-\beta_* / (2\beta_* + t_* - 2)} + \sigma_*/\sqrt{n}$ of the minimax optimal rate.
In particular, he provided some rationale for that the first term $n^{-\beta_* / (2\beta_* + t_* - 2)}$ is sharp.
Furthermore, he constructed a GAN type estimator, which achieves the rate $n^{-\beta_* / (2\beta_* + t_*)} + \sigma_*$.
Therefore, the rate given in Theorem \ref{thm:W-rate} is not optimal.
Nonetheless, the difference is not significant.
Also, the estimator in \cite{chae2022rates} is devised for theoretical purposes, and it is not clear to us how to compute it in practice.
We would like to emphasize that although likelihood-based approaches are not theoretically optimal, they are popularly used in practice because their computation is much easier than that of GAN.

It would also be important to study lower bounds specifically for likelihood approaches considered in this paper.
More specifically, one may try to obtain a sharp lower bound for $\sup_{Q_*} \E W_1(\hat Q_{\widetilde\sigma}, Q_*)$, where $\hat Q_{\widetilde\sigma}$ is a sieve MLE based on the perturbed data $\widetilde \bX_i = \bX_i + \widetilde \bepsilon_i$ with $\widetilde\bepsilon_i \sim \cN(\bzero_D, \widetilde\sigma^2 \Id_D)$, and $Q_*$ ranges over structured distributions considered in Theorem \ref{thm:W-rate}. 
Ideally, we hope 
\bean
	\inf_{\widetilde\sigma \geq 0} \sup_{Q_*} \E W_1(\hat Q_{\widetilde\sigma}, Q_*) \gtrsim n^{-\beta_* / 2(\beta_* + t_*)} + \sigma_*,
\eean
matching with the upper bound given in Theorem \ref{thm:W-rate}.
To achieve this goal, we would need two arguments. 
Each of them is challenging and of independent interest.
Firstly, we would need a sharp lower bound for the approximation error of deep neural networks.
This would be related to \cite{park2021minimum}, but a far more thorough study is necessary.
Another one is regarding the identifiability issue; we would need $\| \bff - \bff_* \| \lesssim W_1( Q_\bff, Q_{\bff_*})$ or a similar inequality, the reverse of $W_1( Q_\bff, Q_{\bff_*}) \lesssim \| \bff - \bff_* \|$.
Obtaining such a reverse inequality is known to be challenging; see \cite{nguyen2013convergence, wei2022convergence}.
Due to these difficulties, we do not consider this problem in this paper and leave it as future work.

\subsection{Effect of $\sigma_*$ into the Convergence Rate}

It is worthwhile to discuss the effect of the noise level $\sigma_*$ into the convergence rate \eqref{eq:rate-final}.
Firstly, suppose that $\sigma_*$ is a fixed positive constant.
Then, the rate \eqref{eq:rate-final} does not give useful information because the right hand side is not small enough.
In fact, estimating $Q_*$ under an additive noise is known as a deconvolution problem, for which extensive studies have been done in the literature (\citealp{fan1991optimal, alexander2009deconvolution, nguyen2013convergence}).
The minimax optimal rate for the Gaussian deconvolution with a fixed $\sigma_*$ is very slow, \eg\ $(\log n)^{-1}$, implying the intrinsic difficulty of the estimation problem.
Such an intrinsic difficulty has also been observed in \cite{genovese2012manifold} who considered a slightly different problem.
Specifically, they obtained the minimax optimal rate for estimating the support of $Q_*$ under the Hausdorff distance, see Theorem 8 therein.
They assumed that $Q_*$ is supported on a low-dimensional manifold, but the intrinsic slow rate $(\log n)^{-1}$ was unavoidable.
Although their manifold estimation problem is slightly different from the deconvolution, they are closely related to each other as discussed in Section 1.1 of \cite{genovese2012manifold}.
Given the inherent challenges of the deconvolution problem, it does not seem possible to achieve a fast convergence rate in estimating $Q_*$ under fixed variance Gaussian noise.
In this sense, the constant variance set-up would not be appropriate for studying the amazing performance of deep generative models theoretically.

The rate \eqref{eq:rate-final} gives meaningful results when $\sigma_*$ is small enough in the sense that $\sigma_*$ converges to zero with a suitable rate as the sample size increases. In this case, data are concentrated in a small neighborhood of a certain low-dimensional structure; hence one may utilize the structural benefit to estimate $Q_*$ efficiently.
Note that although the set-up is not exactly the same as ours and different estimation problems (such as the manifold or regression function) are considered, there are many recent theoretical articles adopting the regime in which data are concentrated around a very small neighborhood of a manifold (\citealp{aamari2018stability, aamari2019nonasymptotic, divol2021minimax, jiao2021deep, puchkin2022structure, berenfeld2022estimating}); see also Remark 4 of \cite{tang2022minimax}.
In these papers, small neighborhoods depend on the sample size and shrink to a low-dimensional manifold.

Despite the above observation, we wish to emphasize that our results  or the probability bounds are again non-asymptotic in nature. That is, for every $n$, our results hold simultaneously  for a range of $\sigma_*$'s  with $\sigma_*\in  [\sigma_{\min}, \sigma_{\max}] $.

\section{Class of True Distributions}
\label{sec:dist}

Asymptotic properties of a sieve MLE are investigated in the previous sections under the assumption that $P_*=P_{\bff_*, \sigma_*}$ for some $\bff_*$ and $\sigma_*$, that is, $P_*$ is the convolution of $Q_{\bff_*}$ and $\cN(\bzero_D, \sigma_*^2 \Id_D)$.
In this  section we characterize  the class of probability distributions of the form $Q_\bff$. In particular, we will show that the class $\{Q_\bff : \bff \in \cF\}$ is quite general to include various structured distributions when $\bff$ ranges over a certain class $\cF$ of structured functions.
Specifically, we will show that various distributions can be represented as $Q_\bff$ for some function $\bff$.
Throughout this section, we assume that $\bZ \sim P_Z$ and $\bY$ is a random vector whose distribution $Q$ satisfies that $Q(\cY) = 1$ for $\cY \subset \bbR^D$.
A primary goal is to find a map $\bff:\cZ \to \bbR^D$ satisfying $Q = Q_\bff$.
\cite{lu2020universal} considered a similar topic, but they did not consider structures of $\bff$ such as the smoothness, which are important for obtaining a fast convergence rate.

\subsection{Case $D=d=1$: 1-dimensional Distributions or Smooth Densities} \label{ssec:dim1}

Suppose that both $\bY$ and $\bZ$ are absolutely continuous real-valued random variables with the cumulative distribution functions $F_Y$ and $F_Z$, respectively.
Then, it is well-known that $F_Y^{-1} ( F_Z(\bZ))$ is distributed as $Q$, where $F_Y^{-1}(u) = \inf\{y \in \bbR: F_Y(y) > u\}$ is the generalized inverse of $F_Y$.
That is, $Q = Q_\bff$, where $\bff = F_Y^{-1} \circ F_Z$.
Furthermore, it is known that the map $\bff$ is the unique optimal transport from $P_Z$ to $Q$ with respect to the quadratic cost function, see Section 2.2 of \cite{villani2003topics}.
If $\bZ$ follows Uniform$(0,1)$, for example, the smoothness of $\bff$ is determined by the smoothness of $F_Y^{-1}$.
Informally, if the pdf $q$ is $\beta$-smooth and strictly positive on $\cZ$, then $F_Y^{-1}$ is $(\beta+1)$-smooth, see Lemma \ref{lem:caffarelli} for a formal statement.
Note that a smooth 1-dimensional function $\bff$ can be approximated by DNN efficiently.
Roughly, if $\bff \in \cH^\beta$, then for any $\epsilon > 0$, there exists $\bff^{\rm nn} \in \cF(L, \bp, s, \infty)$ with $L \asymp \log \epsilon^{-1}$, $|\bp|_\infty \asymp \epsilon^{-1/\beta}$ and $s \asymp \epsilon^{-1/\beta} \log \epsilon^{-1}$ such that $\| |\bff - \bff^{\rm nn}|_\infty \|_\infty \leq \epsilon$, see Theorem 5 of \cite{schmidt2020nonparametric}.

\subsection{Product Distributions}

Assume  that $D = d$ and $\bY = (Y_1, \ldots, Y_D)^T$, where $Y_1, \ldots, Y_D$ are independent random variables.
That is, $Q$ is the product probability of $Q_1, \ldots, Q_D$, where $Q_j$ is the distribution of $Y_j$.
If $Z_1, \ldots, Z_D$ are \iid\ random variables, there exist univariate functions $f_j$, $j=1, \ldots, D$, such that $Q_j$ is the distribution of $f_j(Z_j)$, as argued in Section \ref{ssec:dim1}.
Therefore, the map $\bff$ defined as $\bff(\bz) = (f_1(z_1), \ldots, f_D(z_D))^T$ satisfies that $Q = Q_\bff$.
As before, if densities $q_1, \ldots, q_D$ exist and sufficiently smooth, $\bff$ can be choosen as a smooth function.
Specifically, if each $q_j \in \cH^\beta$ for every $j$, one can find $\bff^{\rm nn} \in \cF(L, \bp, s, \infty)$ with $L \asymp \log \epsilon^{-1}$, $|\bp|_\infty \asymp \epsilon^{-1/\beta}$ and $s \asymp \epsilon^{-1/\beta} \log \epsilon^{-1}$ such that $\| |\bff - \bff^{\rm nn}|_\infty \|_\infty \leq \epsilon$.
That is, we only need to approximate $D$ many 1-dimensional smooth functions.

\subsection{Classical Smooth Densities} \label{ssec:Holder}

Suppose that $D=d$ and $Q$ has the Lebesgue density $q$.
An open set $\Omega \subset \bbR^r$ is said to be uniformly convex if there exists a twice continuously differentiable function $\bh: \bbR^r \to \bbR$ and a constant $\lambda>0$ such that $\Omega = \{\bx \in \bbR^r: \bh(\bx) < 0\}$ and $\nabla^2 \bh(\bx) - \lambda \Id_r$ is positive definite for every $\bx\in \bbR^d$, where $\nabla^2 \bh(\bx)$ is the Hessian matrix.
Note that a uniformly convex set is automatically bounded.
The following lemma is a special case of Theorem 12.50 in \cite{villani2008optimal}, originally proven by \cite{caffarelli1990interior} and \cite{urbas1988regularity}.
As mentioned in \cite{villani2003topics}, techniques involved in Lemma \ref{lem:caffarelli} are really intricate.
We refer to page 139 of \cite{villani2003topics} for more references about this topic.

\begin{lemma} \label{lem:caffarelli}
Suppose that (i) $\cZ$ and $\cY$ are uniformly convex, (ii) $p_Z$ and $q$ are bounded from above and below on $\cZ$ and $\cY$, respectively, and (iii) $q \in \cH^\beta(\cY)$ and $p_Z \in \cH^\beta(\cZ)$ for $\beta > 0$.
Then, there exists a function $\bff = (f_1, \ldots, f_d): \cZ \to \cY$ such that  $Q = Q_\bff$ and $\bff \in \cH^{\beta+1}$.
\end{lemma}

The map $\bff$ in Lemma \ref{lem:caffarelli} is the unique optimal transport from $P_Z$ to $Q$ with respect to the quadratic cost function.
For statistical purpose, a map $\bff$ needs not to be an optimal transport, therefore, conditions on $P_Z$ and $Q$ can be relaxed.
For example, note that the uniform distribution on the unit ball $\cB(\bzero_d)$ has a density which is bounded from above and below, and $\cB(\bzero_d)$ is uniformly convex.
Hence, if $Q$ satisfies the condition in Lemma \ref{lem:caffarelli} and there exists a map $\bh: \cZ \to \cB(\bzero_d)$ such that $\bh(\bZ) \sim {\rm Uniform}(\cB(\bzero_d))$, Lemma \ref{lem:caffarelli} guarantees the existence of $\bff$ satisfying $Q = Q_\bff$.
If $P_Z$ is the uniform distribution on the unit cube $(0,1)^d$, which is a popular choice in practice, such $\bh$ can be chosen as a smooth function, see \cite{harman2010decompositional}.
Conditions on $Q$, such as the uniform convexity of $\cY$, can be relaxed in a similar way.
Finally, we note that if $\bff \in \cH^{\beta+1}$, there exists $\bff^{\rm nn} \in \cF(L, \bp, s, \infty)$ with $L \asymp \log \epsilon^{-1}$, $|\bp|_\infty \asymp \epsilon^{-d/(\beta+1)}$ and $s \asymp \epsilon^{-d/(\beta+1)} \log \epsilon^{-1}$ such that $\| |\bff - \bff^{\rm nn}|_\infty \|_\infty \leq \epsilon$.

\subsection{Distributions on a Manifold}\label{ssec:manifold}

We consider the case where $\cY \subset \bbR^D$ is a topological manifold with dimension $d_* \leq d$.
We start with the case that $\cY$ can be covered by a single chart, that is, there exists a homeomorphism $\bvarphi: \cB_1(\bzero_{d_*}) \to \cY$. 
We further assume that $\bvarphi \in \cH^{\beta+1}$ for $\beta > 0$ as a map from $\cB_1(\bzero_{d_*})$ to $\bbR^D$, and that $\inf_{\bx \in \cB_1(\bzero_{d_*})} |J_{\bvarphi}(\bx)|$ is bounded below by a positive constant, where
\bean
	|J_{\bvarphi}(\bx)| =\sqrt{{\rm det} \left( \frac{\partial \bvarphi}{\partial \bx^T} \frac{\partial \bvarphi}{\partial \bx} \right) }
\eean
is the Jacobian determinant of $\bvarphi$.
Note that a coordinate chart in a smooth manifold is automatically smooth by the definition of a smooth map between manifolds, \cf\ \cite{lee2013smooth}.
Therefore, the ordinary differentiability $\bvarphi \in \cH^{\beta+1}$ is an additional condition.
This kind of condition is frequently used in literature, see \cite{schmidt2019deep, nakada2020adaptive}.

Furthermore, we impose some  smooth conditions on the distribution $Q$.
Note that if $D$ is strictly larger than $d_*$, the distribution $Q$ cannot possess a Lebesgue density because $\cY$ is a null set.
We instead consider a density with respect to the Hausdorff measure.
Let $\cH_{d_*}$ be the $d_*$-dimensional Hausdorff measure in $\bbR^D$, which is normalized so that it is the same as the Lebesgue measure if $D=d_*$.
Suppose that $Q$ allows the Radon--Nikodym derivative $q$ with respect to $\cH_{d_*}$.
We further assume that $q$ is bounded from above and below, and that $q \circ \bvarphi \in \cH^\beta$.
Then, by the change of variable formula, the Lebesgue density of $\widetilde Q$, the distribution of $\bvarphi^{-1}(\bY)$, is given as
\bean
	\widetilde q(\bx) = q\big( \bvarphi(\bx)\big) | J_{\bvarphi}(\bx) |.
\eean
Since $|J_{\bvarphi}(\bx)| \neq 0$ and $\bvarphi \in \cH^{\beta+1}$, it is not difficult to see that $|J_{\bvarphi}(\bx)|$ is bounded from above and below, and the map $\bx \mapsto |J_{\bvarphi}(\bx)|$ belongs to $\cH^\beta$.
Hence, $\widetilde q$ is bounded from above and below, and belongs to $\cH^\beta(\cB_1(\bzero_{d_*}))$.
By Lemma \ref{lem:caffarelli}, under mild assumptions on $P_Z$, there exists $\bg \in \cH^{\beta+1}(\cZ)$ such that $\widetilde Q = Q_\bg$.
Thus, we have $Q = Q_\bff$, where $\bff = \bvarphi \circ \bg \in \cH^{\beta+1}$ is a map from $\cZ$ to $\bbR^D$.
As in Section \ref{ssec:Holder}, one can choose $\bff^{\rm nn} \in \cF(L, \bp, s, \infty)$ with $L \asymp \log \epsilon^{-1}$, $|\bp|_\infty \asymp \epsilon^{-d_*/(\beta+1)}$ and $s \asymp \epsilon^{-d_*/(\beta+1)} \log \epsilon^{-1}$ such that $\| |\bff - \bff^{\rm nn}|_\infty \|_\infty \leq \epsilon$.

Now, we illustrate the case of multiple charts.
Suppose that a distribution $Q$ is supported on a $d_*$-dimensional manifold $\cM$ that can be covered by $J$ charts $(U_j, \bvarphi_j), j=1\ldots, J$, where  $J>1$.
Here, $U_j\subset \cY$  are open sets, with homeomorphism $\bvarphi_j: \cB_1(\bzero_{d_*}) \to U_j$.
As before, we further assume that $\bvarphi_j \in \cH^{\beta+1}$, $\inf_{\bx \in \cB_1(\bzero_{d_*})} |J_{\bvarphi_j}(\bx)|$ is bounded below by a positive constant, $Q$ possesses a Hausdorff density that is bounded from above and below, and that $q \circ \bvarphi_j \in \cH^\beta$.
Let $Q_j(\cdot)=Q(\cdot)/Q(U_j)$ be the normalized measure of $Q$ over $U_j$ and denote its corresponding Hausdorff density as $q_j$.
Note that for $\by\in U_i\cap U_{j}$, one has $q_i(\by)Q(U_i)=q_j(\by)Q(U_j)=q(\by)$ because $Q(U_i)Q_i(\cdot)$ and $Q(U_j)Q_j(\cdot)$ agree with $Q$ on  $U_i\cap U_{j}$. 

Next we will show that $Q$ can be patched together from $Q_j$ via a partition of unity.
Note that a {\it partition of unity} of a topological space $\cY$ is a set  of continuous functions $\{\tau_j: j \in \cJ\}$  from $\cY$ to the unit interval $[0,1]$ such that for every point, $\by\in \cY$, there is a neighborhood $U$ of $\by$ where all but a finite number of the functions are 0, and the sum of all the function values at $y$ is 1, i.e., $\sum_{j\in\cJ}\tau_j (\by)=1$.
A compact manifold $\cM$ always  admits  a \emph{finite partition of unity} $\{\tau_j: \ j=1, \dots, J\}$, $\tau_j(\cdot): \cM\rightarrow [0,1]$ such that 
$\sum _{j=1}^J\tau_j (\by)=1.$
Furthermore, one can construct $\{\tau_j: \ j=1, \dots, J\}$ so that each $\tau_j$ is sufficiently smooth and $\tau_j(\by)=0$ for $\by \notin U_j$, see Lemma 3 of \cite{schmidt2019deep}.

Since $q(\by) = Q(U_j) q_j(\by)$ for each $j$ and $\by\in U_j$, one has $q(\by) = \sum_{j=1}^J Q(U_j) \tau_j(\by) q_j(\by)$.
Let $\widetilde q_j(\by) = c_j \tau_j(\by) q_j(\by)$, where $c_j = [\int \tau_j(\by) dQ_j(\by)]^{-1}$ is the normalizing constant.
Then, $q(\by) = \sum_{j=1}^J \pi_j \widetilde q_j(\by)$, where $\pi_j = Q(U_j) / c_j$.
That is, $q$ is a mixture of $\widetilde q_j$'s.
Since $\widetilde q_j$ is sufficiently smooth, one can construct $\widetilde\bff_j: \widetilde\cZ \to \cY$ such that $\widetilde Q_j$ is the distribution of $\widetilde\bff_j(\widetilde\bZ)$ as in the single chart case, where $\widetilde\cZ$ is a uniformly convex subset of $\bbR^{d_*}$ and $\widetilde\bZ$ follows the uniform distribution on $\widetilde\cZ$.
Let $\cZ = (0,1) \times \widetilde\cZ$ and $P_Z$ be the product distribution of Uniform$(0,1)$ and the distribution of $\widetilde \bZ$.
Let $I_1, \ldots, I_J$ be disjoint consecutive intervals with lengths $\pi_1, \ldots, \pi_J$ partitioning $(0,1)$, that is, $I_1 = (0, \pi_1)$ and $I_j = [\sum_{i=1}^{j-1} \pi_i, \sum_{i=1}^j \pi_i)$ for $j=2, \ldots, J$.
Let $h_j$ be the indicator function for the interval $I_j$.
Then, for a random variable $Z$ following Uniform$(0,1)$, we have $P_Z(h_j(Z) = 1) = 1- P_Z(h_j(Z) = 0) = \pi_j$.
For $\bz = (z_1, \bz_2) \in \bbR^{d_*+1}$, define $\bff(\bz) = \sum_{j=1}^J h_j(z_1) \widetilde\bff_j(\bz_2)$.
Then, it is not difficult to see that $Q = Q_\bff$.
Note that each $\widetilde\bff_j$ can be efficiently approximated by ReLU network functions as the single chart case.
Also, 1-dimensional indicator functions $h_1, \ldots, h_J$ can be approximated by piecewise linear functions.
Therefore, it is easy to approximate them by shallow ReLU network functions.
Finally, the multiplication of $h_j$ and $\widetilde \bff_j$ can also be well-approximated by ReLU networks.

\begin{remark}
Strictly speaking, the regularity of the map $\widetilde\bff_j$ is not guaranteed because $\tau_j$ is not bounded from below.
From the construction of $\tau_j$ in \cite{schmidt2019deep}, however, it can be seen that $\tau_j$ vanishes only at the boundary of $U_j$ (relative to $\cM$).
Hence, one may construct a sufficiently regular $\widetilde\bff_j$ such that $\widetilde Q_j \approx Q_{\widetilde\bff_j}$.
A more rigorous treatment of this topic would be very technical, and we leave it as future work.
\end{remark}

\section{Numerical Experiments}\label{sec:exp}

In this section, we empirically demonstrate that the data perturbation method proposed in Section 3.4  plays an important role to improve the performance of a sieve MLE of deep generative models.
In addition, we illustrate that deep generative models can detect low-dimensional structures well.
Numerical studies are carried out by analyzing various synthetic and real data sets and comparisons are made between our estimators and others such as the MLE of a linear factor model, GAN and Wasserstein GAN. 

\subsection{Synthetic and Real Data Sets}
\subsubsection{Synthetic Data}
\begin{figure}
\bc
	\includegraphics[width=0.8\textwidth]{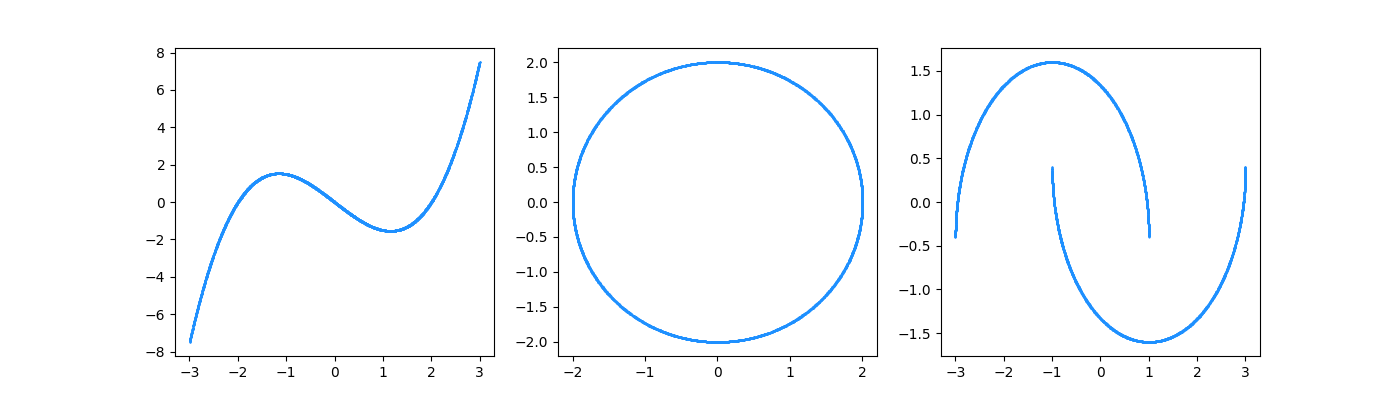}
\ec
\caption{Supports of $Q_*$ for the three synthetic data sets in (\ref{trans_fcs}).
}
\label{fig:synthetic_transformations}
\end{figure}

For simulation study, we firstly consider distributions on 1-dimensional manifolds.
Specifically, we generate data from the model $\bX=\bff_*(\bZ)+\bepsilon_*$ with $D=2$ and  $\sigma_* = 0$, where $\bZ$ is a univariate random variable following Uniform$(0,1)$.
For the true generator $\bff_*=(f_{*1},f_{*2})$, we consider the following three functions:
\begin{align}
\label{trans_fcs}
\begin{array}{cll}
    \text{Case 1.} & f_{*1}(z)=6(z-0.5),& f_{*2}(z)=0.5(z-2)z(z+2)\\
    \text{Case 2.} & f_{*1}(z)=2\cos{(2\pi z)},& f_{*2}(z)=2\sin{(2\pi z)} \\
    \text{Case 3.} & 
    \left\{ 
    \begin{array}{l}
        f_{*1}(z)=2\cos{(2\pi z)}+1,  \\
        f_{*1}(z)=2\cos{(2\pi z)}-1, 
    \end{array}\right.& 
    \begin{array}{ll}
        f_{*2}(z)=2\sin{(2\pi z)}+0.4&\text{ if } z>0.5  \\
        f_{*2}(z)=2\sin{(2\pi z)}-0.4&\text{ otherwise.}
    \end{array}
\end{array}
\end{align}
The supports of $Q_*$ for the three cases are depicted in Figure \ref{fig:synthetic_transformations}.
The generator of Case 2 leads the uniform distribution on a circle.
Note that a circle cannot be covered by a single chart.
Also, for Case 3, the true generator is discontinuous.
In this case, the support of $Q_*$ is the union of two disjoint 1-dimensional manifolds.

We next consider two more distributions, a distribution on the Swiss roll (\citealp{marsland2015machine})   and the uniform distribution on  the  sphere, which are supported on 2-dimensional manifolds with the ambient space $\bbR^3$.
The distribution on the Swiss roll  is the distribution of $\bff_*(\bZ)$, where $\bZ$ follows the uniform distribution on $(0,1)^2$ and the true generator $\bff_* = (f_{*1}, f_{*2}, f_{*3}): (0,1)^2 \to \bbR^3$ is defined as
\bean
	t_1 = 1.5\pi(1+2 z_1), && \quad
	t_2 = 21 z_2,
	\\
	f_{*1}(z_1, z_2) = t_1 \cos(t_1), && \quad
	f_{*2}(z_1, z_2) = t_2, \quad
	f_{*3}(z_1, z_2) = t_1 \sin(t_1).
\eean
Similar to the circle, the sphere cannot be covered by a single chart.
In all the experiments, the sample sizes of validation and test data are set to be 3,000, while the  training sample size varies.

\subsubsection{Big Five Personality Traits Data Set}
The big five personality traits data set (Big-five; \cite{goldberg1990alternative}) consists of answers for 50 questions, with the five-level Likert scale (1 to 5) from 1,015,342 respondents.
This data set has been frequently analyzed in literature with linear factor models, see \cite{ohn2021posterior} and references therein.
We only use the data of the 874,434 respondents who answer to all questions completely. 
Each variable is rescaled to take values from $-1$ to 1.
We randomly draw 20,000 samples from the entire data, 10,000 of which are used as validation data and the others as test data. 
The remains  are used as training data.

\subsubsection{MNIST and Omniglot Data Sets}
We analyze two well-known image data sets, MNIST and Omniglot.
MNIST data set (\citealp{lecun1998gradient}) contains handwritten digit images of $28\times 28$ pixel sizes and 
has a training data set consisting of 60,000 images and a test data set of 10,000 images. 
We randomly sample 10,000 images from the training data set and use them as validation data. 

Omniglot (\citealp{lake2015human}) data set consists of various character images of $28\times 28$ pixel sizes taken from 50 different alphabets. 
It has 24,345 training samples and 8,070 test samples. 
As before, we split the training data set into two subsets, each of which has 20,000 and 4,345 samples, respectively, and use one for training data and the other for validation data. 

\subsection{Learning Algorithm to Obtain the  MLE} \label{ssec:algorithm}

Assume that the generator $\bff = \bff_\theta$ is parametrized by $\theta$.
With a slight abuse of notation, let $p_{\theta, \sigma} = p_{\bff_\theta, \sigma}$, that is,
\begin{equation*}
	p_{\theta, \sigma}(\bx)=\int \phi_{\sigma}\left(\bx-\bff_\theta(\bz)\right)dP_Z(\bz).
\end{equation*}
Mostly, the log-likelihood is computationally intractable.
Alternatively, one  can maximize a lower bound of the log-likelihood by use of a family of variational distributions using methods of variational inference (\citealp{jordan1999introduction}).
The most well-known algorithm is the variational autoencoder (VAE; \citealp{kingma2013auto,rezende2014stochastic}) and the lower bound used in VAE is often called the ELBO (evidence lower bound).

Various alternative  lower bounds of the log-likelihood that are tighter than the ELBO  but still computationally tractable, have  been proposed afterwards, see \cite{burda2016importance, cremer2017reinterpreting, kingma2016improving, rezende2015variational, salimans2015markov, sonderby2016ladder}. Among these, the importance weighted autoencoders (IWAE, \citealp{burda2016importance}) is an important variant of the VAE.
Recently, it is shown that IWAE can be understood as an EM algorithm to obtain the MLE, see \cite{dieng2019reweighted, kim2020casting}.
Thus, we use the IWAE algorithm to obtain a sieve MLE.
Specifically, let $\bz \mapsto q_\phi(\bz\mid\bx)$ be a variational density parametrized by $\phi$.
A popular choice for $q_\phi(\cdot \mid \bx)$ is the density of $\cN(\bmu_\phi(\bx), \bSigma_\phi(\bx))$, where $\bx \mapsto \bmu_\phi(\bx)$ and $\bx \mapsto \bSigma_\phi(\bx)$ are DNN functions with network parameters $\phi$.
For given \iid\ samples $\bZ_1, \ldots, \bZ_K$ from $q_\phi(\cdot|\bx)$, let 
\bean\label{iwae-appr}
	\hat{L}^{\text{IWAE}}(\theta,\phi,\sigma;\bx):=\log \left( \frac{1}{K} \sum_{k=1}^K \frac{p_{\theta,\sigma}(\bx,\bZ_k)}{q_\phi(\bZ_k|\bx)} \right),
\eean
where $p_{\theta,\sigma}(\bx, \bz) = p_Z(\bz) \phi_\sigma(\bx-\bff_\theta(\bz))$ and $K$ is a given positive integer.
Then, IWAE simultaneously estimates $\theta,\sigma$ and $\phi$ by maximizing $\sum_{i=1}^n \hat{L}^{\text{IWAE}}(\theta,\phi,\sigma;\bX_i).$
We set $K=10$ throughout our experiments. 

\subsection{Implementation Details} \label{ssec:implementation}
\subsubsection{Data Perturbation} 
The model is trained after perturbing the training data by an artificial noise $\widetilde{\bepsilon}\sim\cN(\bzero_D,\widetilde{\sigma}^2\Id_D)$.
For each data set, we consider various values of $\widetilde{\sigma}$.

\subsubsection{Architectures} 
For analyzing five synthetic and Big-five data sets, we consider DNN architectures with the leaky ReLU activation function (\citealp{xu2015empirical}).
For the variational distribution $q_\phi(\cdot\mid \bx)$, we use the multivariate normal distribution $\cN(\bmu_{\phi}(\bx),\bSigma_{\phi}(\bx))$, where $\bSigma_{\phi}(\bx)$ is a diagonal matrix.
Both the mean $\bmu_{\phi}$ and variance $\bSigma_{\phi}$ are modelled by  DNNs.
For synthetic data, we set $L=2$, $d=10$, $\bp = (d, 200, 200, D)$ for $\bff_\theta$, and $L=2$, $\bp=(D,200,200,d)$ for $\bmu_{\phi}$ and $\bSigma_{\phi}$.
For the Big-five data set, we set $L=3$, $d=5$, $\bp = (d, 200, 200, 200, D)$ for $\bff_\theta$, and $L=3$, $\bp = (D, 200, 200, 200, d)$ for $\bmu_{\phi}$ and $\bSigma_\phi$. 

For analyzing two image data, we use a deep convolutional neural network (\citealp{radford2015unsupervised}) with $L=6$ and the ReLU activation function for modeling $\bff_{\theta}$.  
Also, convolutional neural networks with $L=6$ and the leaky ReLU activation function are used to build model architectures for $\bmu_{\phi}$ and $\bSigma_{\phi}$.
For the both data sets, we set $d=40$.

\subsubsection{Optimization}
We train deep generative models using the Adam optimization algorithm (\citealp{kingma2014adam}) with a mini-batch size of 100. 
The learning rate is fixed as $10^{-3}$ for synthetic and Big-five data, and $3\times 10^{-4}$ for two image data.

\subsubsection{Sparse Learning Framework} 
For learning sparse generative models, we adopt the pruning algorithm proposed by \cite{han2015learning}. 
Firstly, a non-sparse model is trained with a pre-specified maximum number of training epochs, 200 in our experiments, and then the number of training epochs which minimizes the IWAE loss on the validation data is chosen. Next, the model is pruned by zeroing out small weights.
Specifically, 25\% of small weights are replaced by zero.
We then re-train the model keeping the zero weights unchanged.
This procedure is repeated one more time to make 50\% of the total weights become zero in the final model.

\subsection{Performance Comparisons}

The performance of a given estimator $\hat Q$ is evaluated by the Wasserstein distance $W_1(\hat Q, Q_*)$  estimated on test data as follows.
Let $\hat\bbQ_M$ be the empirical measure based on the $M$ \iid\ samples from $\hat Q$.
Note that it is easy to generate samples from $\hat Q$ via the estimated generator.
Similarly, let $\bbQ_{M*}$ be the empirical measure based on the $M$ observations in test data.
Then, $W_1(\hat Q, Q_*)$ can be estimated by $W_1(\hat \bbQ_M, \bbQ_{M*})$.
In general, $W_1(\hat \bbQ_M, \bbQ_{M*})$ can be computed via a linear programming.
We use a more stable algorithm developed by \cite{cuturi2013sinkhorn}.
We call $W_1(\hat \bbQ_M, \bbQ_{M*})$ the {\it estimated $W_1$ distance}.

\subsubsection{Results for Synthetic Data} 
\begin{figure}
\bc
	\includegraphics[width=0.32\textwidth]{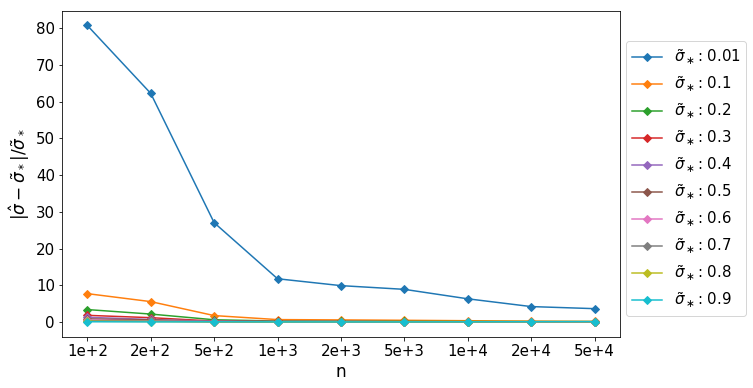}
	\includegraphics[width=0.32\textwidth]{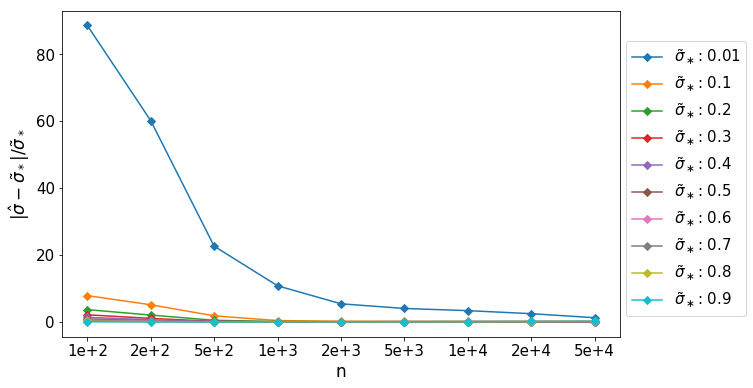}
	\includegraphics[width=0.32\textwidth]{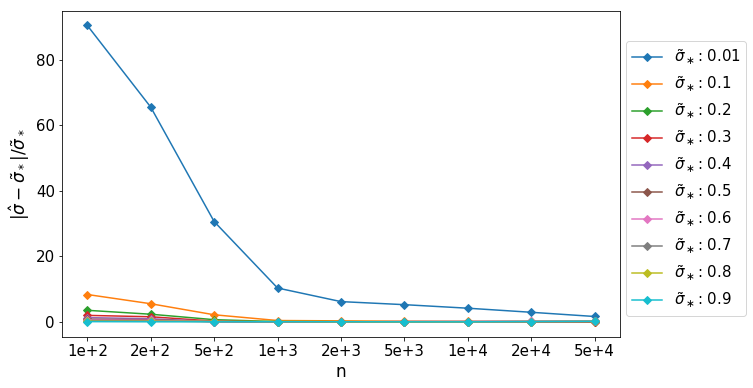}
\ec
\caption{Values of $|\hat{\sigma}-\widetilde{\sigma}_*|/\widetilde{\sigma}_*$ for various $\widetilde{\sigma}_*$ and $n$ for the three 
1-dimensional synthetic data sets.} \label{fig:syn_results_sigdiff}
\end{figure}

For the three 1-dimensional synthetic data sets, various training sample sizes ranging from 100 to 50,000 are considered . 
For each case, we obtain a sieve MLE for three times with random initialization and 
report the average based on the three sieve MLEs.
Firstly, we trace the estimated variance $\hat{\sigma}^2.$
Figure \ref{fig:syn_results_sigdiff} draws the values of $|\hat\sigma - \widetilde\sigma_*| / \widetilde\sigma_*$ as the sample size increases, where $\widetilde{\sigma}_*^2=\sigma_*^2+\widetilde{\sigma}^2 = \widetilde\sigma^2$. 
It seems  that $|\hat\sigma - \widetilde\sigma_*| / \widetilde\sigma_* \to 0$ as $n$ increases
regardless of the value of $\widetilde{\sigma}_*^2$, which
suggests that  sieve MLEs perform reasonably well.

\begin{figure}
\bc
	\includegraphics[width=0.31\textwidth]{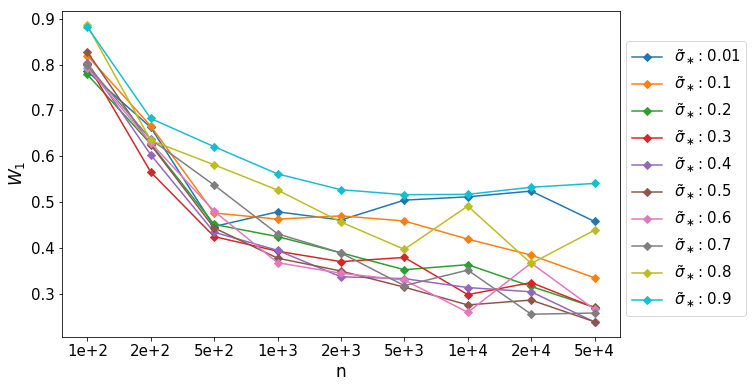}
	\hspace{.1cm}
	\includegraphics[width=0.31\textwidth]{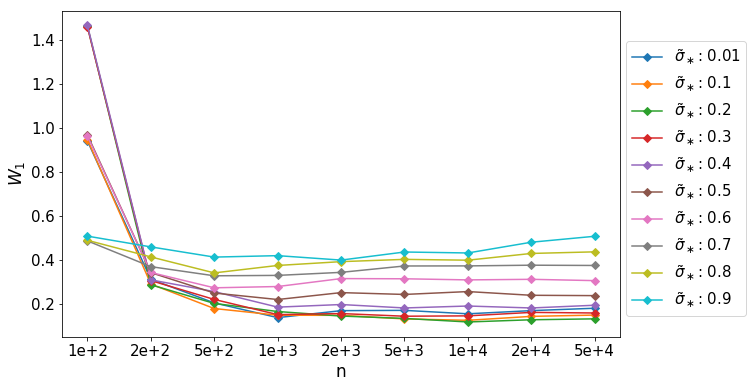}
	\hspace{.1cm}
	\includegraphics[width=0.31\textwidth]{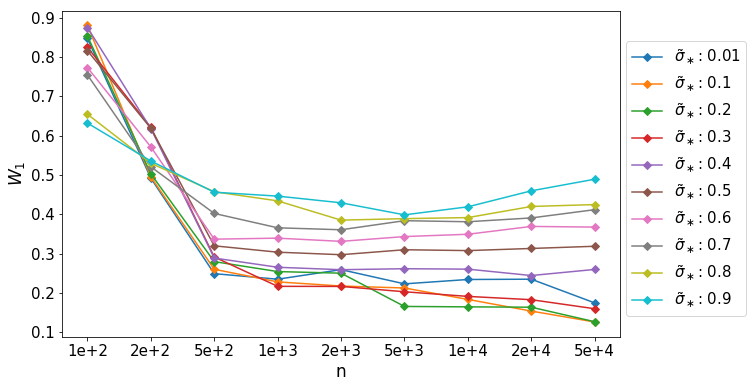}
\ec
\caption{The estimated $W_1$ distance over the sample size with various values of $\widetilde{\sigma}$
for the three 1-dimensional synthetic data sets.} \label{fig:syn_results}
\end{figure}

\begin{figure}
\bc
	\includegraphics[width=0.3\textwidth]{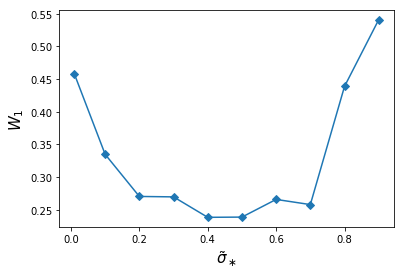}
	\includegraphics[width=0.3\textwidth]{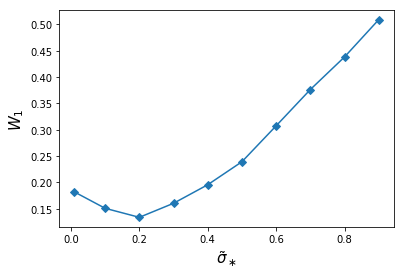}
	\includegraphics[width=0.3\textwidth]{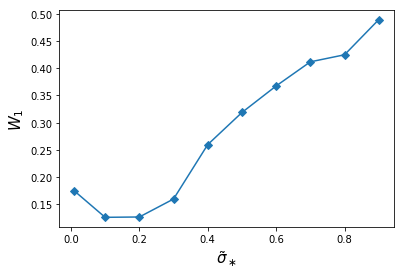}
\ec
\caption{The estimated $W_1$ distance over $\widetilde{\sigma}$ with the training sample size being fixed at $n=50,000$
 for the three 1-dimensional synthetic data sets .} \label{fig:syn_results2}
\end{figure}

\begin{figure}
\bc
	\includegraphics[width=0.26\textwidth]{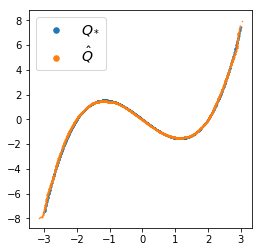}
	\includegraphics[width=0.26\textwidth]{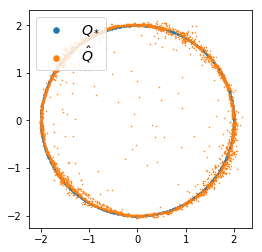}
	\includegraphics[width=0.26\textwidth]{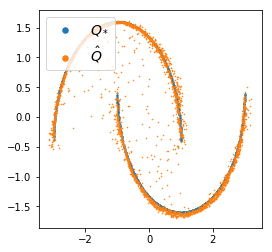}
\ec
\caption{Generated samples from $\hat Q$ for the three 1-dimensional synthetic data sets.}
\label{fig:synthetic1_generation}
\end{figure}

The estimated $W_1$ distances for various training sample sizes are shown in Figure \ref{fig:syn_results}.
It is interesting to see that the estimated $W_1$ distance of a sieve MLE does not converge to 0 when
$\widetilde{\sigma}^2$ is either too small or too large, which well corresponds to Theorem \ref{thm:rate-W}.
Figure \ref{fig:syn_results2} provides the curves of the estimated $W_1$ distances 
over the degree of perturbation (i.e. $\widetilde{\sigma}$) with the training sample size being fixed at $n=50,000.$
As can be seen, the estimated $W_1$ distance is minimized at an intermediate value of $\widetilde\sigma$ in all three cases, which again confirms the validity of our theoretical results.
Figure \ref{fig:synthetic1_generation} presents generated samples from $\hat Q$
estimated with $n=50,000$ and the optimal choice of $\widetilde\sigma$ that minimizes the estimated $W_1$ distance. 

\begin{figure}
\bc
	\includegraphics[width=0.3\textwidth]{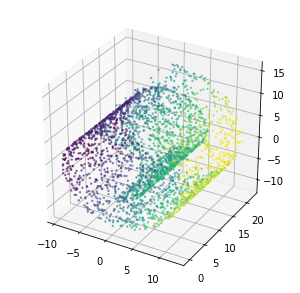}
	\includegraphics[width=0.3\textwidth]{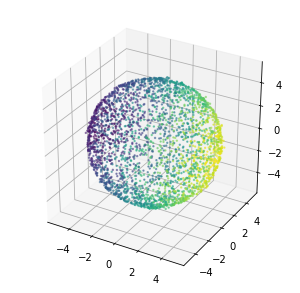}
\ec
\caption{Generated samples from $\hat Q$ for the two 2-dimensional synthetic data sets.}
\label{fig:synthetic2_generation}
\end{figure}

Similar phenomena can be found for the Swiss roll and sphere models. 
That is, the estimated $W_1$ distance is minimized at an intermediate value of $\widetilde\sigma$.
Generated samples from $\hat Q$ with $n=50,000$ and the optimal choice of $\widetilde\sigma$
are plotted over the  support of $Q_*$ in Figure \ref{fig:synthetic2_generation}.

\subsubsection{Results for  Big-five Data Set}

\begin{figure}
\bc
	\includegraphics[width=0.3\textwidth]{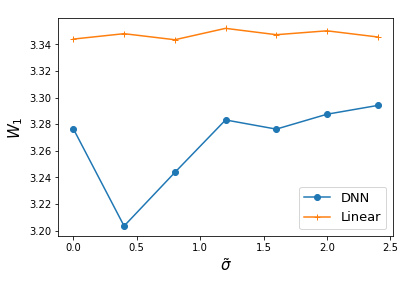}
	\includegraphics[width=0.3\textwidth]{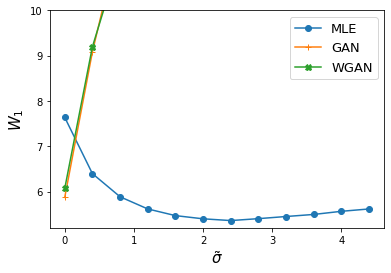}
	\includegraphics[width=0.3\textwidth]{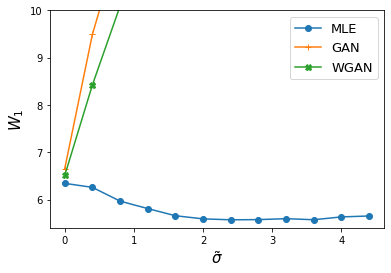}
\ec
\caption{The estimated $W_1$ distance over $\widetilde{\sigma}$ for Big-five (left), MNIST (middle) and Omniglot (right) data.}\label{fig:image_results_comp}
\end{figure}

The Big-five data set is trained with various values of $\widetilde \sigma,$ and
the estimated $W_1$ distances over various values of $\widetilde\sigma$ are depicted in the left panel of Figure \ref{fig:image_results_comp}.
Again, it is clear that the estimated $W_1$ distance is minimized at an intermediate value of $\widetilde\sigma.$
In addition, we provide the results of the MLE of a sparse linear factor model for comparison, which has been considered in literature for analysing the Big-five data set, see \cite{ohn2021posterior}.
A deep generative model is significantly better than a sparse linear factor model, which indicates
that nonlinear factor models are necessary for practical data analysis.

\begin{figure}
\bc
	\includegraphics[width=0.35\textwidth]{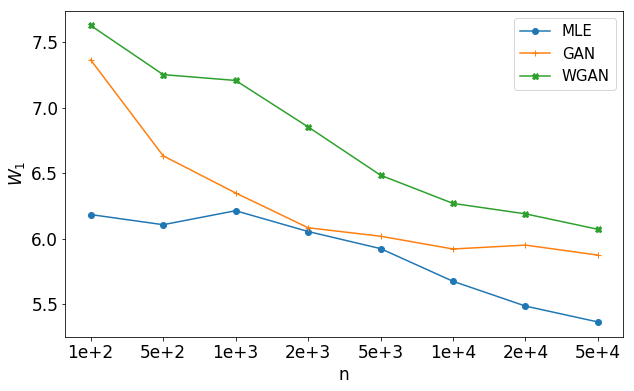}
	\includegraphics[width=0.35\textwidth]{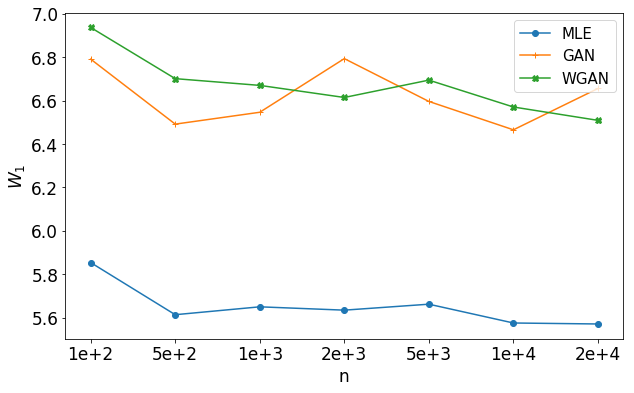}
\ec
\caption{The estimated $W_1$ distance over the sample size for MNIST (lefg) and Omniglot (right) data. An optimal $\widetilde\sigma$ is chosen for sieve MLEs based on the validation errir, and no data perturbation is applied for GAN and WGAN.}\label{fig:image_results_comp2}
\end{figure}

\subsubsection{Results for MNIST and Omniglot Data Sets}

The results about the estimated $W_1$ distance for various $\widetilde\sigma$ are shown in the middle and right panels of Figure \ref{fig:image_results_comp}. 
Again, we observe that the estimated $W_1$ distance is minimized at an intermediate value of $\widetilde\sigma.$ 
On the other hand, the data perturbation does not work at all for GAN and Wasserstein GAN.
Moreover, a sieve MLE with proper data perturbation outperforms GAN and Wasserstein GAN for the both image data sets, as detailed in Figure \ref{fig:image_results_comp2}.

\begin{figure}
\bc
	\includegraphics[width=0.22\textwidth]{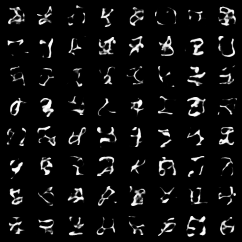}
	\includegraphics[width=0.22\textwidth]{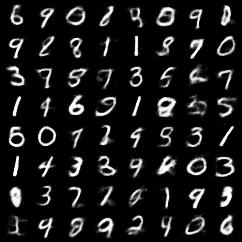}
	\includegraphics[width=0.22\textwidth]{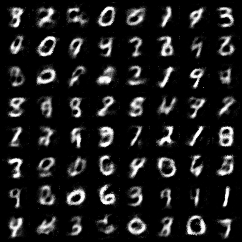}
	\\ \vspace{.1cm}
	\includegraphics[width=0.22\textwidth]{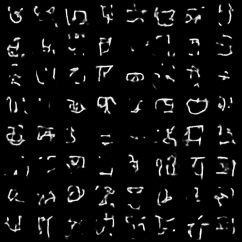}
	\includegraphics[width=0.22\textwidth]{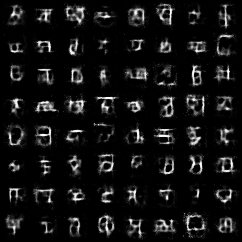}
	\includegraphics[width=0.22\textwidth]{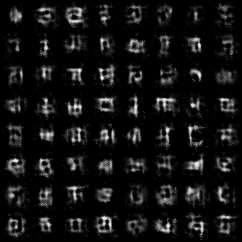}
\ec
\caption{Randomly generated images from a sieve MLE $\hat Q$ for MNIST (upper) and Omniglot (lower). We considered three values of $\widetilde{\sigma}$, 0.0, 2.0 and 4.0 from left to right.} \label{fig:gen_samples}
\end{figure}

\begin{figure}
\bc
	\includegraphics[width=0.22\textwidth]{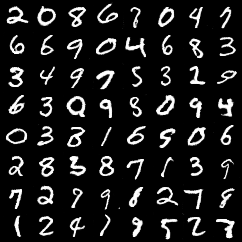}
	\includegraphics[width=0.22\textwidth]{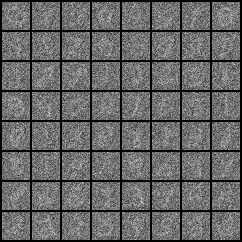}
	\includegraphics[width=0.22\textwidth]{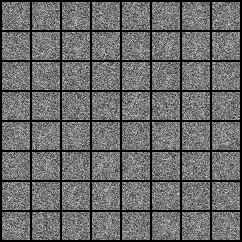}
	\\ \vspace{.1cm}
	\includegraphics[width=0.22\textwidth]{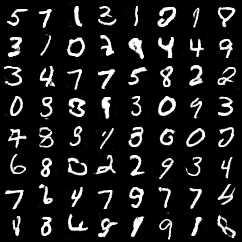}
	\includegraphics[width=0.22\textwidth]{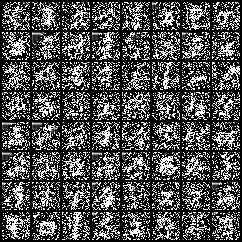}
	\includegraphics[width=0.22\textwidth]{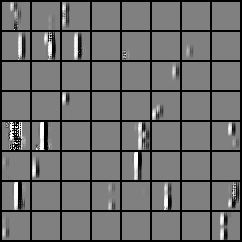}
\ec
\caption{Randomly generated images by GAN (upper) and WGAN (lower) estimators for MNIST. We consider three values of $\widetilde{\sigma}$, 0.0, 2.0 and 4.0 from left to right.} \label{fig:gen_samples_baselines_mnist}
\end{figure}

\begin{figure}
\bc
	\includegraphics[width=0.22\textwidth]{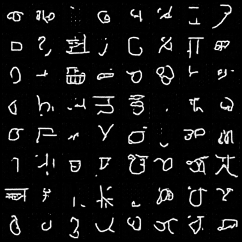}
	\includegraphics[width=0.22\textwidth]{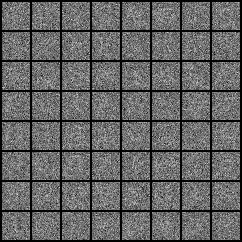}
	\includegraphics[width=0.22\textwidth]{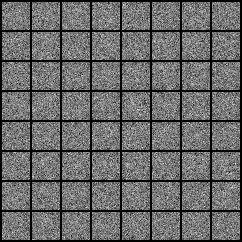}
	\\ \vspace{.1cm}
	\includegraphics[width=0.22\textwidth]{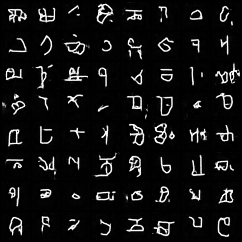}
	\includegraphics[width=0.22\textwidth]{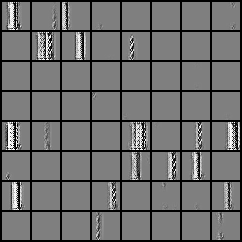}
	\includegraphics[width=0.22\textwidth]{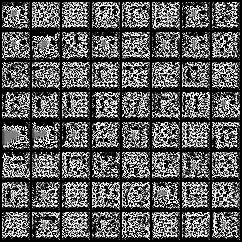}
\ec
\caption{Randomly generated images by GAN (upper) and WGAN (lower) estimators  for Omniglot. We consider three values of $\widetilde{\sigma}$, 0.0, 2.0 and 4.0 from left to right.}\label{fig:gen_samples_baselines_omniglot}
\end{figure}

Figure \ref{fig:gen_samples} presents randomly generated images from  sieve MLEs $\hat Q$ for MNIST and Omniglot data sets with three values of $\widetilde\sigma$, 0.0, 2.0 and 4.0.
It is obvious that $\widetilde\sigma=2.0$ gives the best results for the both data, which implies that 
the estimated $W_1$ distance is positively related to the cleanness of corresponding synthetic images.
Randomly generated images of GAN and Wasserstein GAN learned with data perturbation
for MNIST and Omniglot are given in Figures \ref{fig:gen_samples_baselines_mnist} and \ref{fig:gen_samples_baselines_omniglot}, respectively, which again confirms that
data perturbation is not helpful for GAN and Wasserstien GAN to generate synthetic images.

\subsection{Meta-learning for Low-dimensional Composite Structures}

In Section 3.2, we have proved that a sieve MLE of deep generative models
can capture a low-dimensional composition structure well. Using this flexibility of a sieve MLE, we can learn
a low-dimensional composite structure from a sieve MLE as follows.
For example, suppose that $\bff_*$ possesses a generalized additive model (GAM) structure such as
$$f_{*j}(\bz)=g_{*j1}(z_1)+\cdots + g_{*j d}(z_d)$$
for $j=1,\ldots,D.$ Then, we can estimate the component functions $g_{*jl}, l=1,\ldots,d$ by minimizing
$$\sum_{i=1}^N \left( \hat{f}_j(\bz_i)-g_{j1}(z_{i1})+\cdots + g_{j d}(z_{id})\right)^2$$
under certain regularity conditions, where $\bz_i$'s are independently generated samples from $P_Z.$

We investigate the above meta-modeling approach
by simulation. We generate data of size 50,000 from the following two generative models:
\medskip

Model 1: GAM
\begin{eqnarray*}
&&\bz=(z_1,z_2,z_3)\sim \mathcal{N}(0,\Id_3)\\
&&f_{*1}(\bz) = -2.3+\frac{1}{0.7+\exp(0.3-2z_1)}+0.3 z_2^2\\
&&f_{*2}(\bz) = 0.9 + 0.8z_1-0.1z_1^3+\log(z_2^2+1.5)-0.4z_3^2\\
&&f_{*3}(\bz) = 1.8+\frac{3.5}{2z_2^2+z_2+4}-0.2\exp(z_3)\\
&&f_{*4}(\bz) = 1.2z_1-0.1z_2^3+0.05z_3^4\\
&&f_{*5}(\bz) = 3+0.5\log(2.5+\exp(z_1))-0.2\exp(z_3+0.2)
\end{eqnarray*}
\medskip

Model 2: Non-additive model
\begin{eqnarray*}
&&\bz=(z_1,z_2,z_3)\sim \mathcal{N}(0,\Id_3)\\
&&f_{*1}(\bz) = \frac{5z_3}{3.7+\exp(-2z_1+0.4z_2)}\\
&&f_{*2}(\bz) = 0.9 -0.1z_1-0.2z_1(z_2-0.1)^2+0.15z_1z_3\\
&&f_{*3}(\bz) = \log(2+(z_1-z_2)^2)-0.2z_1\exp(0.2*z_3)\\
&&f_{*4}(\bz) = 1.5-0.3z_1^2+0.07z_1z_2z_3\\
&&f_{*5}(\bz) = \frac{3z_1-1.2}{z_2^2+2z_2+3.3}+0.5\log(1+(z_1-0.1)^2+z_2^2z_3^2)
\end{eqnarray*}
\medskip

\begin{figure}
\bc
	\includegraphics[width=0.3\textwidth]{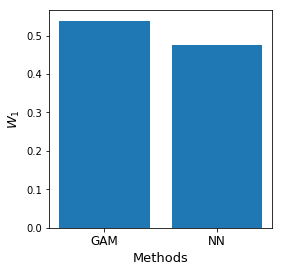}
	\includegraphics[width=0.3\textwidth]{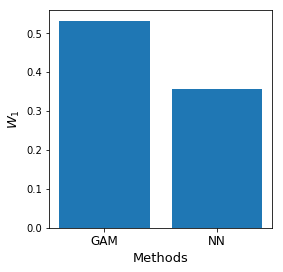}
\ec
\caption{Estimated $W_1$ distances of a sieve MLE and the estimated GAM for Model 1 (left) and
Model 2 (right)}
\label{fig:composite-sim}
\end{figure}

We estimated the components of the GAM from a sieve MLE of the deep generative model
by the proposed meta-modeling and compare
the estimated $W_1$ distances of the original sieve MLE and the estimated GAM in Figure \ref{fig:composite-sim}. 
The orginal sieve MLE outperforms the GAM for the two simulation models but
the difference of the estimated $W_1$ distances
is smaller for the first model where the true model is a GAM than the second model, which indicates
that the sieve MLE captures the underlying low-dimensional composite structure well.

\begin{figure}
\bc
	\includegraphics[width=0.3\textwidth]{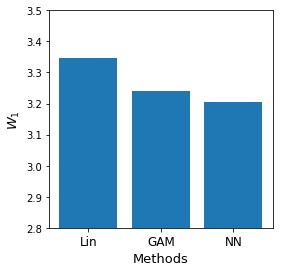}
	\includegraphics[width=0.3\textwidth]{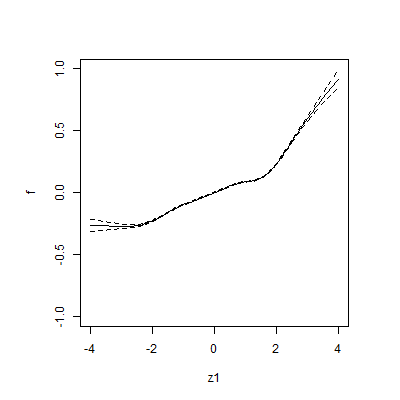}
	\includegraphics[width=0.3\textwidth]{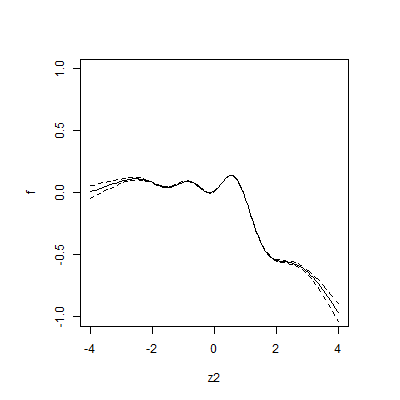}
	\includegraphics[width=0.3\textwidth]{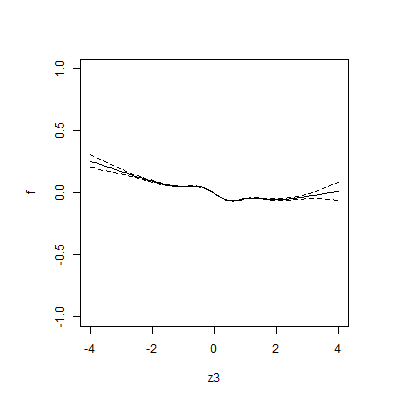}
	\includegraphics[width=0.3\textwidth]{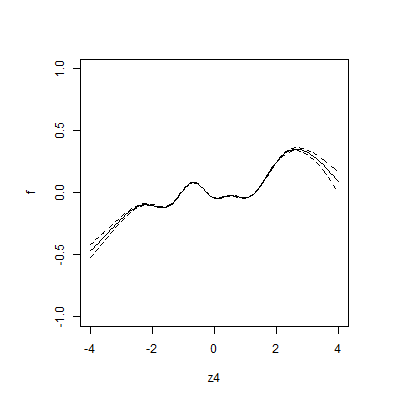}
	\includegraphics[width=0.3\textwidth]{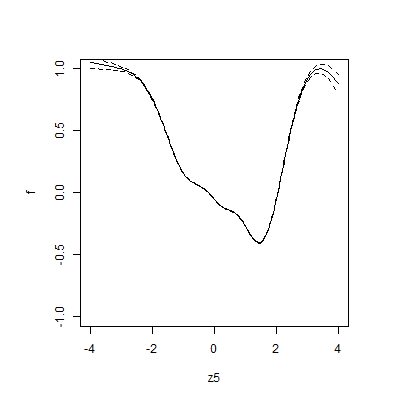}
\ec
\caption{The estimated $W_1$ distances of (sieve) MLEs of the linear model, deep generative model and 
the estimated GAM (upper left) and the five estimated component functions of a randomly selected corrdinate (i.e. $\hat{f}_{14}$) 
of the GAM for the Big-five data set} \label{fig:gam}
\end{figure}

For the Big-five data set, the upper left panel of Figure \ref{fig:gam} compares the estimated $W_1$ distances
of three estimates, (sieve) MLEs of the linear and deep generative models and
the estimated GAM obtained by the meta-learning.
The GAM improves over the linear model but 
is slightly inferior to the deep generative model.
 The five estimated component functions for $\hat{f}_{14},$ a randomly selected coordinate,
are drawn in Figure \ref{fig:gam}. 
Some of them clearly show non-linearity, which partly explains why the performance of the deep generative model is much better than the linear factor model. 

\section{Discussion}\label{sec:conclusion}

\begin{table}
\centering
{\small\begin{tabular}{c|llll}
	\hline\hline
	 & Manifold & Noise level & Upper bound & Lower bound 
	\\ \hline\hline
	G1 & $\cC^2$ & $|\bepsilon_*|_\infty \lesssim 1$ & $n^{-2/(2+d_*)}$ & $n^{-2/(2+d_*)}$
	\\ \hline
	G2 & $\cC^2$ & $ \bepsilon_* \sim N(\bzero_D, \Id_D)$ & $(\log n)^{-1/2}$ & $(\log n)^{-1}$
	\\ \hline
	P & $\cC^2$ & $|\bepsilon_*|_\infty \leq \sigma_* \lesssim n^{-2/(3d_*+8)}$ & $n^{-2/d_*} \vee (\sigma_*^2/n)^{2/(d_*+4)}$ & $(\sigma_*^2/n)^{2/(d_*+4)}$
	\\ \hline
	A & $\cC^\alpha$ & $|\bepsilon_*|_\infty \leq \sigma_* \lesssim n^{-1/d_*}$ & $n^{-\alpha/d_*} \vee \sigma_*$ & $n^{-\alpha/d_*} \vee (\sigma_*/n)^{\alpha/(d_*+\alpha)}$
	\\ \hline
	D & $\cC^2$ & $|\bepsilon_*|_\infty \lesssim n^{-2/d_*}$ & $n^{-2/d_*}$ & $n^{-2/d_*}$
	\\ \hline\hline
\end{tabular}}
\caption{\label{tab:manifold} Convergence rates of the manifold estimators
 with respect to the Hausdorff distance from existing papers: \cite{genovese2012minimax} (G1), \cite{genovese2012manifold} (G2), \cite{puchkin2022structure} (P), \cite{aamari2019nonasymptotic} (A), \cite{divol2021minimax} (D). $\cC^\alpha$ in the second column refers that $\cM_*$ is a differentiable manifold of order $\alpha$. For \cite{genovese2012minimax} and \cite{aamari2019nonasymptotic}, it is assumed that $\bepsilon_*$ is perpendicular to the manifold, see \cite{genovese2012minimax} for details.}
\end{table}

In this work, we consider the estimation of a distribution of high-dimensional data based on a deep generative model which includes the estimation of classical smooth densities and distributions supported on lower-dimensional manifolds as special cases.  The case when $Q_*$ is supported on a smooth manifold $\cM_*$ with dim$(\cM_*)=d_*$, is the most interesting and challenging case.
For this model, one may be interested in estimating the manifold or the support of $\cM_*$ itself.
One can easily construct an estimator for $\cM_*$ by $\hat\cM = \hat\bff(\cZ)$ based on an estimator $\hat\bff$.
The performance of $\hat\cM$ might be evaluated through a convergence rate with respect to the Hausdorff metric.
Some existing results  on convergence rates are summarized in Table \ref{tab:manifold} with assumptions on the underlying manifold and noise level.
All these papers assume that the reach of the underlying manifold is bounded below by a positive constant.
Technical assumptions from different papers may vary, but none of these papers explicitly consider the regularity of $q_*$, the density with respect to the volume measure.
In particular, \cite{genovese2012minimax} assumed that the error vector is perpendicular to the manifold which is somewhat a strong condition. 
In \cite{genovese2012manifold}, the perpendicular error is replaced by standard Gaussian error leading to a slow convergence rate.
This slow rate is standard in a deconvolution problem with a supersmooth Gaussian kernel.
The other three papers considered bounded errors which decay to zero with suitable rates.
If the noise level is sufficiently small and $\cM_* \in \cC^2$, the minimax convergence rate would be $n^{-2/d_*}$.
It would be interesting to investigate whether an estimator $\hat\cM$ constructed from a deep generative model can achieve this rate.
More generally, it would be worthwhile  to study the manifold estimation problem through the lens of deep generative models.

We have some interesting observations from the results of analysis of the two image data sets in Section 5.
While GAN and WGAN generate clearer images than a sieve MLE, the performance of a sieve MLE in terms of the evaluation metric $W_1(\hat \bbQ_M, \bbQ_{M*})$ is better than both, if a suitable degree of perturbation is applied.
Surprisingly, opposite results are obtained if FID (\Frechet\ Inception distance; \cite{heusel2017gans}) is used as a measure of performance.
Note that FID is an approximation of $L^2$-Wasserstein distance in the feature space of Inception model (\citealp{szegedy2016rethinking}), and it is one of the most popularly used performance measures in image generation problems.
The obtained FID values are 2.76, 4.19 and 9.58 for GAN, WGAN and sieve MLE with the optimal $\widetilde\sigma$, respectively.
That is, both GAN and WGAN are significantly better than a sieve MLE in terms of FID.
At this point, we are not aware of any reason why two performance measures, $W_1(\hat \bbQ_M, \bbQ_{M*})$ and FID, yield opposite results, which we leave as a future work.


\acks{The authors are very grateful to the Editor,  the Associate Editor and the  reviewers for their valuable comments  which have led to substantial improvement in the paper. 
MC was supported by Samsung Science and Technology Foundation under Project Number SSTF-BA2101-03. 
DK was supported by the National Research Foundation of Korea (NRF) grant funded by the Korea government (MSIT) (No. NRF-2022R1G1A1010894).
YK was supported by the NRF grant (No. 2020R1A2C3A01003550) and Institute of Information \& communications Technology Planning \& Evaluation (IITP) grant (No. 2022-0-00184, Development and Study of AI Technologies to Inexpensively Conform to Evolving Policy on Ethics) funded by MSIT.
LL would like to acknowledge the generous support of NSF grants DMS CAREER 1654579  and DMS 2113642. }

\appendix

\section{Proofs}

\subsection{Proof of Lemma \ref{lem:bracket-entropy}}

For $\bff_1, \bff_2 \in \cF$ with $\| |\bff_1 - \bff_2|_\infty \|_\infty \leq \eta_1$, we have that
\bean
	&& p_{\bff_1, \sigma}(\bx) - p_{\bff_2, \sigma}(\bx)
	= \int \phi_\sigma(\bx - \bff_1(\bz)) \left\{ 1 - \frac{\phi_\sigma(\bx - \bff_2(\bz))}{\phi_\sigma(\bx - \bff_1(\bz))}\right\} dP_Z(\bz)
	\\
	&& = \int \phi_\sigma(\bx - \bff_1(\bz)) \left[ 1 - \exp\left\{ \frac{|\bx - \bff_1(\bz)|_2^2 - |\bx - \bff_2(\bz)|_2^2}{2\sigma^2} \right\} \right] dP_Z(\bz)
	\\
	&& \leq \int \phi_\sigma(\bx - \bff_1(\bz)) \frac{|\bx - \bff_2(\bz)|_2^2 - |\bx - \bff_1(\bz)|_2^2}{2\sigma^2} dP_Z(\bz)
	\\
	&& = \int \phi_\sigma(\bx - \bff_1(\bz)) \frac{|\bff_2(\bz)|_2^2 - |\bff_1(\bz)|_2^2 - 2\bx^T(\bff_2(\bz) - \bff_1(\bz)) }{2\sigma^2} dP_Z(\bz)
	\\
	&& \leq \int \phi_\sigma(\bx - \bff_1(\bz)) \frac{KD\eta_1 + \sqrt{D} |\bx|_2 \eta_1}{\sigma^2} dP_Z(\bz),
\eean
where the last inequality holds because $||\bff_1(\bz)|_2^2 - |\bff_2(\bz)|_2^2| \leq 2KD \eta_1$ and $|\bx^T(\bff_1(\bz) - \bff_2(\bz))| \leq \sqrt{D}|\bx|_2 \eta_1$.
Since $|\bx|_2 \leq |\bx-\bff(\bz)|_2 + |\bff(\bz)|_2 \leq 1 + |\bx-\bff(\bz)|_2^2 + \sqrt{D}K$ and $|\bx|_2^2 \phi_\sigma(\bx) /(2\sigma^2) \leq (2\pi\sigma^2)^{-D/2} / e$, the last display is further bounded by
\be\begin{split}\label{eq:ff-bound}
	& \eta_1 \int \phi_\sigma(\bx - \bff_1(\bz)) \left( \frac{2KD + \sqrt{D}}{\sigma^2} + \frac{\sqrt{D} |\bx-\bff_1(\bz)|_2^2}{\sigma^2} \right) dP_Z(\bz)
	\\
	& \leq \eta_1 \left(2\pi\sigma^2\right)^{-D/2} \left( \frac{2KD + \sqrt{D}}{\sigma^2} + \frac{2\sqrt{D}}{e} \right).
\end{split}\ee
Also, for $\sigma_1, \sigma_2 \in [\sigma_{\min}, \sigma_{\max}]$ with $|\sigma_1 - \sigma_2| \leq \eta_2$, it holds that $|\sigma_1^{-2} - \sigma_2^{-2}| \leq \sigma_1^{-2} \sigma_2^{-2} (\sigma_1+\sigma_2) \eta_2$ and $|\log (\sigma_2/\sigma_1)| \leq \eta_2 / (\sigma_1 \wedge\sigma_2)$.
Hence
\be\begin{split}\label{eq:ss-bound}
	&p_{\bff, \sigma_1}(\bx) - p_{\bff, \sigma_2}(\bx)
	\\
	&= \int \phi_{\sigma_1}(\bx - \bff(\bz)) \left[ 1 - \left(\frac{\sigma_1}{\sigma_2}\right)^D \exp\left\{ \frac{|\bx - \bff(\bz)|_2^2}{2} \left( \frac{1}{\sigma_1^2} - \frac{1}{\sigma_2^2} \right) \right\} \right] dP_Z(\bz)
	\\
	&\leq \int \phi_{\sigma_1}(\bx - \bff(\bz)) \left\{\frac{|\bx - \bff(\bz)|_2^2}{2} \left( \frac{1}{\sigma_2^2} - \frac{1}{\sigma_1^2} \right) - D\log\frac{\sigma_1}{\sigma_2} \right\} dP_Z(\bz)
	\\
	&\leq \eta_2 \int \phi_{\sigma_1}(\bx - \bff(\bz)) \left( \frac{ (\sigma_1+\sigma_2) |\bx - \bff(\bz)|_2^2}{2 \sigma_1^2 \sigma_2^2} + \frac{D}{\sigma_1 \wedge \sigma_2} \right)  dP_Z(\bz)
	\\
	&\leq \eta_2 \left(2\pi \sigma_1^2\right)^{-D/2} \left( \frac{\sigma_1 + \sigma_2}{e \sigma_2^2} + \frac{D}{\sigma_1 \wedge \sigma_2} \right).
\end{split}\ee

Let $\epsilon > 0$ be given.
Let $\{\bff_1, \ldots, \bff_{N_1}\}$ and $\{\sigma_1, \ldots, \sigma_{N_2}\}$ be  $\eta_1$-covering of $\cF$ and $\eta_2$-covering of $[\sigma_{\min}, \sigma_{\max}]$, respectively.
By \eqref{eq:ff-bound} and \eqref{eq:ss-bound}, there exist constants $c_1=c_1(D, K)$ and $c_2=c(D)$ such that $\eta_1 = c_1 \sigma_{\min}^{D+2} \epsilon$ and $\eta_2 = c_2 \sigma_{\min}^{D+1} \epsilon$ implies that $\{p_{\bff_i, \sigma_j}: i=1, \ldots, N_1, j=1, \ldots, N_2\}$ forms an $\epsilon/2$-covering of $\cP$ with respect to $\|\cdot\|_\infty$.
For each $(i,j)$, define $l_{ij}$ and $u_{ij}$ as
\bean
	l_{ij}(\bx) = \max\{p_{\bff_i, \sigma_j}(\bx) - \epsilon/2, 0\}
	\quad{\rm and} \quad
	u_{ij}(\bx) = \min\{p_{\bff_i, \sigma_j}(\bx) + \epsilon/2, H(\bx)\},
\eean
where $H(\bx) = \sup_{p \in \cP} p(\bx)$ is an envelop function of $\cP$.
Note that
\bean
	&& H(\bx) \leq \left( 2\pi\sigma_{\min}^2 \right)^{-D/2} \sup_{|\by|_\infty\leq K} \exp\left( - \frac{|\bx-\by|_2^2}{2\sigma_{\max}^2} \right)
	\\
	&& \leq \left( 2\pi\sigma_{\min}^2 \right)^{-D/2} \exp\left( - \frac{|\bx|_2^2 - 2K^2D}{4\sigma_{\max}^2} \right)
	= 2^{D/2} \left(\frac{\sigma_{\max}}{\sigma_{\min}}\right)^D e^{K^2D/2} \phi_{\sqrt{2}\sigma_{\max}}(\bx),
\eean
where the second inequality holds because $|\bx-\by|_2^2 \geq |\bx|_2^2/2 - |\by|_2^2 \geq |\bx|_2^2/2 - K^2D$.
Since $\int_{|\bx|_\infty > t} \phi_\sigma(\bx) d\bx \leq D e^{-t^2/(2\sigma^2)}$, we have that $\int_{|\bx|_\infty > B} H(\bx) d\bx \leq \epsilon$, 
where 
\bean
	B = 2 \sigma_{\max} \left( \log \frac{1}{\epsilon} + D \log \frac{\sigma_{\max}}{\sigma_{\min}} + \frac{D}{2}\log 2 + \frac{K^2D}{2} + \log D \right)^{1/2}.
\eean
It follows that
\bean
	\int \{u_{ij}(\bx) - l_{ij}(\bx)\} d\bx 
	&\leq& \int_{|\bx|_\infty \leq B} \epsilon~ d\bx + \int_{|\bx|_\infty > B} H(\bx) d\bx 
	\leq \left((2B)^D+1 \right) \epsilon \defeq \delta^2.
\eean
Since $d_H^2(u_{ij}, l_{ij}) \leq \|u_{ij}-l_{ij}\|_1$, we have that 
\bean
	N_{[]}(\delta, \cP, d_H) \leq N_{[]}(\delta^2, \cP, \|\cdot\|_1) \leq N_1 N_2
	\leq  \frac{\sigma_{\max}-\sigma_{\min}}{\eta_2} N(\eta_1, \cF, \| |\cdot|_\infty\|_\infty).
\eean
Since $\epsilon (\log \epsilon^{-1})^{D/2} \leq \sqrt{\epsilon}$ for every small enough $\epsilon$, once $\delta$ is small enough, say $\delta \leq \delta_*$ for some $\delta_* = \delta_*(D)$, it holds that $\epsilon \geq c_3 \delta^4 \{\log(\sigma_{\max}/\sigma_{\min})\}^{-D}$, where $c_3 = c_3(D, K, \sigma_{\max})$.
Hence,
\bean
	\eta_1 \geq \frac{c_1 c_3 \sigma_{\min}^{D+3} \delta^4}{\sigma_{\min} \{\log(\sigma_{\max}/\sigma_{\min})\}^D}.
\eean
Since $\sigma_{\min} \leq 1$, $\sigma_{\min} \{\log(\sigma_{\max}/\sigma_{\min})\}^D$ is bounded by a constant which depends only on $\sigma_{\max}$ and $D$, so $\eta_1$ is bounded below by $c_4 \sigma_{\min}^{D+3} \delta^4$, where $c_4 = c_4(D, K, \sigma_{\max})$.
A similar lower bound can be obtained for $\eta_2$, which completes the proof.

\subsection{Proof of Theorem \ref{thm:rate-general}}

We will apply Theorem 4 of \cite{wong1995probability} with $\alpha=0+$.
Choose four absolute constants $c_1, \ldots, c_4$ as in their Theorem 1.
These constants can be chosen so that $c_1 = 1/3$ and $c_3 > 2$.
Define $c$ and $c'$ as in the statement of Lemma \ref{lem:bracket-entropy}.

For every $\delta \in (0, c_3 \delta_*]$,
\bean
	\log N_{[]} (\delta/c_3, \cP, d_H) 
	\leq 4(s+1) \log \delta^{-1} + s A + (D+3)(s+1) \log \sigma_{\min}^{-1} + c_5 s
\eean
by Lemma \ref{lem:bracket-entropy}, where $c_5=c_5(c, c', c_3)$.
Hence,
\bean
	&& \int_{\epsilon^2/2^8}^{\sqrt{2} \epsilon} \sqrt{\log N_{[]} (\delta/c_3, \cP, d_H)}~d\delta
	\\
	&& \leq \sqrt{2}\epsilon \sqrt{s A + (D+3)(s+1) \log \sigma_{\min}^{-1} + c_5 s }
	+ \sqrt{2} \epsilon \sqrt{4(s+1)} \sqrt{\log \frac{2^8}{\epsilon^2}}
\eean
for every $\epsilon \leq \sqrt{2} \delta_* \leq c_3 \delta_* / \sqrt{2}$.
For $\epsilon=\epsilon_n=c_6 \sqrt{n^{-1} s \{ A + \log ( n/\sigma_{\min})\}}$ with a large enough constant $c_6=c_6(c_4, c_5, D)$, the last display is bounded by $c_4 n^{1/2} \epsilon_n^2$ for every $n$, so Eq.~(3.1) of \cite{wong1995probability} is satisfied.
Note that  Eq.~(3.1) of \cite{wong1995probability} still holds if $c_6$ is replaced by any constant larger than $c_6$.

It is well-known (see Example B.12 of \cite{ghosal2017fundamentals}) that
\bean
	K(p_*, p_{\bff, \sigma_*}) 
	&\leq& \int K\Big(N\big(\bff_*(\bz), \sigma_*^2\big), N\big(\bff(\bz), \sigma_*^2\big)\Big) dP_Z(\bz)
	\\
	&=& \int \frac{|\bff_*(\bz) - \bff(\bz)|_2^2}{2\sigma_*^2} dP_Z(\bz)
	\leq \frac{D \delta_{\rm app}^2}{2\sigma_*^2} \defeq \delta_n.
\eean
Also, it is easy to see that
\bean
	\int \left(\log \frac{\phi_{\sigma}(\bx)}{\phi_{\sigma}(\bx-\by)} \right)^2 \phi_{\sigma}(\bx) d\bx
	= \int \frac{|\by|_2^4 + 4|\bx^T \by|^2}{4\sigma^2} \phi_{\sigma}(\bx) d\bx
	\leq \frac{|\by|_2^4}{4\sigma^2} + |\by|_2^2 \int \frac{|\bx|_2^2}{\sigma^2} \phi_{\sigma}(\bx) d\bx.
\eean
Combining this with Example B.12, (B.17) and Exercise B.8 of \cite{ghosal2017fundamentals}, we have that
\bean
	&&\int \left(\log \frac{p_*(\bx)}{p_{\bff, \sigma_*}(\bx)} \right)^2 dP_*(\bx)
	\\
	&& \leq	\iint \left(\log \frac{\phi_{\sigma}(\bx - \bff_*(\bz))}{\phi_{\sigma}(\bx-\bff(\bz))} \right)^2 \phi_{\sigma}(\bx - \bff_*(\bz)) d\bx dP_Z(\bz) + 4K(p_*, p_{\bff, \sigma_*})
	\\
	&&\leq \frac{D^2 \delta_{\rm app}^4}{4\sigma_*^2} + D \delta_{\rm app}^2 \int \frac{|\bx|_2^2}{\sigma_*^2} \phi_{\sigma_*}(\bx) d\bx + \frac{2D \delta_{\rm app}^2}{\sigma_*^2}
	\leq c_7 \frac{\delta_{\rm app}^2}{\sigma_*^2} \defeq \tau_n,
\eean
where $c_7 = c_7(D)$.
(Note that both $\delta_n$ and $\tau_n$ need not depend on $n$. We use the notations $\delta_n$ and $\tau_n$ for the notational consistency with Theorem 4 of \cite{wong1995probability}).
Let $\epsilon_n^* = \epsilon_n \vee \sqrt{12\delta_n}$.
Then, Theorem 4 of \cite{wong1995probability} implies that
\bean
	P_* \Big( d_H(\hat p, p_*) > \epsilon_n^* \Big) 
	\leq 5 e^{-c_2 n \epsilon_n^{*2}} + \frac{12 \tau_n}{n\epsilon_n^{*2}} 
	\leq 5 e^{-c_2 n \epsilon_n^{*2}} + \frac{\tau_n}{n\delta_n} 
	= 5 e^{-c_2 n \epsilon_n^{*2}} + \frac{2c_7^2}{Dn}
\eean
By re-defining constants, the proof is complete.

\subsection{Proof of Corollary \ref{cor:composition}}
By Lemma 5 of \cite{schmidt2020nonparametric}, we have 
\bean
	\log N(\delta, \cF, \| |\cdot|_\infty\|_\infty) \leq s\{c_4 (\log n)^2 +  \log \delta^{-1}\}
\eean
for every $\delta > 0$, where $c_4 = c_4(q, \bd, \bt, \bbeta, K)$.
By applying Lemma \ref{lem:composition-approx} and Theorem \ref{thm:rate-general} with $A = c_4 (\log n)^2$, we have the conclusion.

\subsection{Proof of Theorem \ref{thm:rate-W}}

For any constant $c_0 = c_0(D, K, r_*)$, if $\epsilon + \sigma_* \sqrt{ \log \epsilon^{-1}} \geq c_0$, the assertion of Theorem \ref{thm:rate-W} holds trivially by taking a large enough constant $C = C(D, K, r_*)$.
Therefore, it suffices to prove the assertion of Theorem \ref{thm:rate-W} when $\epsilon$ and $\sigma_* \sqrt{\log \epsilon^{-1}}$ are sufficiently small.

For given $\epsilon \in (0,1]$, suppose that $d_H(p_{\bff, \sigma}, p_*) \leq \epsilon$ and $\| |\bff|_\infty\|_\infty \leq K$.
Throughout this proof, $P_{\bff, \sigma}$ and $Q_\bff$ will be denoted as $P$ and $Q$, respectively.
Let $\bY, \bY_*, \bepsilon, \bepsilon_*$ be independent random vectors, with the underlying probability $\nu$ such that $\bY \sim Q$, $\bY_* \sim Q_*$, $\bepsilon \sim \cN(\bzero_D, \sigma^2 \Id_D)$, $\bepsilon_* \sim \cN(\bzero_D, \sigma_*^2 \Id_D)$.

Since
\bean
	\int_{|\bx|_2 > t} \phi_\sigma(\bx) d\bx \leq \int_{|\bx|_\infty > t / \sqrt{D}} \phi_\sigma(\bx) dx \leq D e^{-t^2/(2D\sigma^2)}
\eean
for any $t > 0$, we have $\int_{|\bx|_2 > t_*} \phi_{\sigma_*}(\bx) d\bx \leq \epsilon$ with $t_* = (2D\sigma_*^2 \log (D/\epsilon))^{1/2}$.
Hence,
\bean
	1 - P_*\left(\cM_*^{t_*} \right) = \nu\left( \bY_* + \bepsilon_* \notin \cM_*^{t_*} \right)
	\leq \nu( |\bepsilon_*|_2 > t_*) \leq \epsilon.
\eean
Since $|P(B) - P_*(B)| \leq d_H(P, P_*) \leq \epsilon$ for every Borel set $B$, see Eq. (8) of  \cite{gibbs2002choosing}, we have that $P(\cM_*^{t_*}) \geq 1- 2\epsilon$.

We will next prove that $\sigma \leq 2t_*$, which is the main part of the proof.
For this, we assume on the contrary that $\sigma > 2t_*$ which we will show lead to a contraction. Firstly,  if $\sigma > r_*/2$, then $1 - P([-K-t_*,K+t_*]^D)$ is bounded below by a constant that depends on $K, D$ and $r_*$, which contradicts to $P(\cM_*^{t_*}) \geq 1- 2\epsilon$ provided that $t_*$ and $\epsilon$ are smaller than a certain threshold depending only on $K, D$ and $r_*$.
(Note that $t_*$ and $\epsilon$ are sufficiently small as assumed at the beginning of the proof.)
If $\sigma \in [2t_*, r_*/2]$, then we claim that for every $\bx \in \bbR^D$, there exists $\by\in\bbR^D$ such that $|\bx-\by|_2 \leq \sigma$ and $\cB_{\sigma/2}(\by) \cap \cM_*^{t_*} = \emptyset$.
Let $\rho(\bx, \cM_*) = \inf\{|\bx - \bx'|_2: \bx' \in \cM_*\}$.
The proof of the claim is divided into three cases.

(Case 1) $\rho(\bx, \cM_*) \geq \sigma$: Obviously, one can choose $\by=\bx$.

(Case 2) $\rho(\bx, \cM_*) \in (0, \sigma)$: Let $\bx_0$ be the unique Euclidean projection of $\bx$ onto $\cM_*$, and $\bx_t = \bx_0 + t(\bx-\bx_0)$.
Define two continuous functions $d_0(t) = |\bx_t - \bx_0|_2$ and $d(t) = \rho(\bx_t, \cM_*)$.
Note that $d_0(t) = d(t)$ for all $t \in [0,1]$.
Otherwise, $|\bx_t - \bz|_2 < |\bx_t - \bx_0|_2$ for some $t \in[0,1]$ and $\bz\in\cM_*\backslash{\bx_0}$.
Since $\bx_t$ lies in the line segment with end points $\bx$ and $\bx_0$,
\bean
	|\bx-\bx_0|_2 = |\bx-\bx_t|_2 + |\bx_t - \bx_0|_2 > |\bx-\bx_t|_2 + |\bx_t - \bz|_2 \geq |\bx-\bz|_2,
\eean
and thus, $\bx_0$ cannot be the unique projection of $\bx$ onto $\cM_*$.
Note also that $d(t) = d_0(t)$ for all $t \in [1, 1+ \sigma/|\bx-\bx_0|_2]$.
Otherwise, $\{ t\in [1, 1+ \sigma/|\bx-\bx_0|_2]: d(t) < d_0(t)\}$ is a non-empty set with the infimum $t_0$, and it is not difficult to see that $\bx_{t_0}$ has at least two Euclidean projection onto $\cM_*$.
Let $\by = \bx_{1+\sigma/|\bx-\bx_0|_2}$.
Then, we have $|\by-\bx|_2 = \sigma$ and $\rho(\by, \cM_*) = |\by - \bx_0|_2 = |\bx-\bx_0|_2+\sigma$.
Since $t_* \leq \sigma/2$, we have $\cB_{\sigma/2}(\by) \cap \cM_*^{t_*} = \emptyset$

(Case 3) $\rho(\bx, \cM_*)=0$: Since $\cB_\delta(\bx)$ is not contained in $\cM_*$ for any $\delta > 0$, one can choose $\bx' \in \cB_\delta(\bx) \backslash \cM_*$.
If $\delta$ is small enough, by Case 2, there exists $\by'$ such that $|\bx' - \by'|_2 \leq \sigma$ and $\cB_{\sigma/2}(\by') \cap \cM_*^{t_*} = \emptyset$.
Note that $|\bx - \by'|_2 \leq |\bx-\bx'|_2 + |\bx'-\by'|_2 \leq \delta + \sigma$.
One can take $\by$ as any limit point of $\by'$ as $\delta \to 0$.

By the claim, we have
\bean
	\nu \Big( \bY + \bepsilon\notin \cM_*^{t_*} \mid \bY = \bx \Big)
	\geq \nu \Big( \bepsilon \in \cB_{\sigma/2}(\by-\bx) \Big)
\eean
for every $\bx \in \bbR^D$.
Since $|\by-\bx|_2 \leq \sigma$, the right hand side is bounded below by a positive constant, say $c$, that depends only on $D$.
It follows that $P(\cM_*^{t_*}) = \nu(\bY + \bepsilon \in \cM_*^{t_*}) \leq 1-c$, which contradicts $P(\cM_*^{t_*}) \geq 1- 2\epsilon$ for small enough $\epsilon$.
This completes the proof of $\sigma \leq 2t_*$.

Note that the $\ell_1$-diameter of $[-K, K]^D$ is $2KD$, $W_1 \leq W_2$ and $W_1$ is bounded by a multiple of the total variation, see Theorem 4 of \cite{gibbs2002choosing}.
Also, it is easy to see that $W_2(P_*, Q_*) \leq \sigma_*$ and $W_2(P, Q) \leq \sigma$.
Hence, 
\bean
	W_1(Q_*, Q) \leq W_2(Q_*, P_*) + W_1(P_*, P) +  W_2(P, Q)
	\leq  \sigma_* + KD \|p - p_*\|_1 +  \sigma.
\eean
Since $\|p - p_*\|_1 \leq 2 d_H(p, p_*)$ and $\sigma \leq 2t_*$, the proof is complete.

\subsection{Proof of Theorem \ref{thm:W-rate}}
Let $\widetilde p_* = p_{\bff_*, \widetilde\sigma_*}$, where $\widetilde\sigma_* = \sigma_* + n^{-\beta_*/\{2(\beta_*+ t_*)\}}$.
Also, let $\widetilde\alpha = -\log \widetilde\sigma_* / \log n$, that is, $\widetilde\sigma_* = n^{-\widetilde\alpha}$.
Then, by Corollary \ref{cor:composition}, \eqref{eq:rate-general} holds with
\bean
	\epsilon_n^* = C n^{-\frac{\beta_* - t_* \widetilde\alpha}{2\beta_* + t_*}} (\log n)^{3/2},
\eean
where $C = C(q, \bd, \bt, \bbeta, K, D, \sigma_{\max}, \gamma)$.

Firstly, suppose that $\alpha < \beta_*/ \{2(\beta_* + t_*)\}$.
In this case, $\sigma_* < \widetilde\sigma_* \leq 2\sigma_*$, so
\bean
	\alpha - \frac{\log 2}{\log n} \leq \widetilde\alpha < \alpha.
\eean
Hence, $\epsilon_n^*$ can be re-written as
\bean
	\epsilon_n^* = C' n^{-\frac{\beta_* - t_* \alpha}{2\beta_* + t_*}} (\log n)^{3/2}
\eean
with an adjusted constant $C' = C'(q, \bd, \bt, \bbeta, K, D, \sigma_{\max}, \alpha, \gamma)$ satisfying $2^{-t_* / (2\beta_* + t_*)} C \leq C' < C$.

Similarly, if $\alpha \geq \beta_*/ \{2(\beta_* + t_*)\}$, we have 
\bean
	\frac{\beta_*}{2(\beta_* + t_*)} - \frac{\log 2}{\log n} \leq \widetilde\alpha \leq \frac{\beta_*}{2(\beta_* + t_*)}.
\eean
Hence, $\epsilon_n^*$ can be re-written as
\bean
	\epsilon_n^* = C'' n^{-\frac{\beta_*}{2(\beta_* + t_*)}} (\log n)^{3/2}
\eean
with $C'' = C''(q, \bd, \bt, \bbeta, K, D, \sigma_{\max}, \alpha, \gamma)$.

Finally, Theorem \ref{thm:rate-W} gives the desired result with re-defined constants.

\vskip 0.2in
\bibliography{bib-short}

\end{document}